\documentclass[journal]{IEEEtran}
\usepackage{amsmath,amsfonts}
\usepackage{array}
\usepackage{textcomp}
\usepackage{stfloats}
\usepackage{url}
\usepackage[pdftex]{graphicx}
\usepackage{amsmath}
\usepackage{eqparbox}
\usepackage{graphicx}
\usepackage{hyperref}
\usepackage{caption}
\usepackage{graphicx} 
\usepackage{threeparttable}
\usepackage[caption=false,font=footnotesize]{subfig}
\usepackage{url}
\usepackage{xcolor}
\usepackage{breakurl}
\usepackage{multirow}
\usepackage[ruled,vlined]{algorithm2e}
\usepackage{verbatim}
\usepackage{graphicx}
\def\BibTeX{{\rm B\kern-.05em{\sc i\kern-.025em b}\kern-.08em
    T\kern-.1667em\lower.7ex\hbox{E}\kern-.125emX}}
\usepackage{flushend}
\begin{document}
\title{An Improved Adaptive PID Optimizer with Enhanced Convergence and Stability for Deep Learning}

\author{
	\IEEEauthorblockN{Saurabh Saini\IEEEauthorrefmark{1},
		Kapil Ahuja\IEEEauthorrefmark{1}\IEEEauthorrefmark{2}$^{\S}$,
		Thomas Wick\IEEEauthorrefmark{2},
		Saurav Kumar\IEEEauthorrefmark{3}
	}
	
	\IEEEauthorblockA{\IEEEauthorrefmark{1}
		Department of Computer Science \& Engineering, Indian Institute of Technology Indore, India.}
	
	\IEEEauthorblockA{\IEEEauthorrefmark{2}
		Leibniz Universit\"at Hannover,
		Institut f\"ur Angewandte Mathematik, Germany.}
	
	\IEEEauthorblockA{\IEEEauthorrefmark{3}
		National Remote Sensing Centre, Indian Space Research Organisation, India.}
	
	\thanks{$^{\S}$Corresponding author: Kapil Ahuja (kahuja@iiti.ac.in)}
	\thanks{S. Saini: phd2101101005@iiti.ac.in; T. Wick: thomas.wick@ifam.uni-hannover.de; S. Kumar: saurav.nrsc@gmail.com}
}

\maketitle

\begin{abstract}
Optimization is essential in deep learning. The foundational method upon which most optimizers are built is momentum-based stochastic gradient descent. However, it suffers from two key drawbacks. \textit{First}, it has  noisy and varying gradients, and \textit{second}, it has an overshoot phenomenon. To address noisy gradients, Adam was proposed, which remains the most widely used adaptive optimizer. To address the overshoot phenomenon, a control-theory-based PID optimizer was proposed. To tackle both the limitations within a single framework, several variants of Adaptive PID (AdaPID) have recently been proposed.

Although AdaPID performs well, it still inherits two critical drawbacks from Adam, namely convergence and stability issues. In this work, we address both these limitations. To fix the convergence issue, we uniquely integrate the idea of using a non-increasing effective learning rate into AdaPID (originally proposed in AMSGrad, an extension of Adam). To fix the stability issue, we innovatively integrate a gradient difference based modulation factor into AdaPID (originally proposed in DiffGrad, another extension of Adam). Combining both these ideas in AdaPID, results in our novel IAdaPID-ADG optimizer.

We evaluate our proposed optimizer on multiple datasets, including benchmark datasets (MNIST and CIFAR10) and real-world datasets (IARC and AnnoCerv). The IAdaPID-ADG substantially outperforms all competing optimizers. Additionally, we perform an ablation study on the MNIST dataset to demonstrate the contribution of each added component.

\end{abstract}


\section{Introduction}
Deep learning has emerged as a transformative paradigm across a wide range of domains, including computer vision \cite{krizhevsky2012imagenet, lan2017learning}, image processing \cite{dong2015image, chen2017deeplab}, signal processing \cite{yu2010deep, zhang2012deep}, robotics \cite{mnih2015human}, and natural language processing \cite{greff2016lstm, collobert2008unified}. This success is largely driven by the availability of large-scale datasets and powerful GPU-based computational resources. At the heart of every deep learning system lies the training process. From an optimization perspective, training amounts to minimizing a loss function over a high-dimensional parameter space. The choice of optimizer directly governs its speed, stability, and quality of convergence, and therefore design of efficient and reliable optimization algorithms is a fundamental challenge in deep learning research.

Stochastic Gradient Descent (SGD) is the classical foundation of modern deep learning optimizers \cite{robbn1951stochastic}. Despite its simplicity and strong generalization properties, SGD suffers from slow convergence in large-scale settings. This is because it relies solely on the current gradient, without using any historical information. To address this, momentum was introduced, which accumulates past gradients to accelerate and stabilize training \cite{polyak1964some, nesterov1983method}.

However, momentum-based SGD still has two key drawbacks. First is noisy and varying gradients across parameters. A single global learning rate cannot handle this well, as different parameters may require very different step sizes during training. Adam \cite{kingma2014adam} addressed this by assigning an adaptive learning rate to each parameter. 

The second is the overshoot phenomenon, where accumulated momentum drives parameters past the optimal point, causing a back-and-forth movement around the optimum. The PID optimizer \cite{wang2020pid} addressed this by regulate parameter updates via proportional, integral, and derivative terms, inspired by classical control theory.

To tackle both the limitations at once, multiple variants of AdaPID \cite{weng2022adapid, dai2023pid, jian2025adapid} have been proposed, combining the adaptive learning rate idea with the PID control structure. Despite its strengths, AdaPID inherits two critical drawbacks from Adam, namely convergence and stability issues. We address both with the following contributions:
\begin{itemize}
	\item[(a).] To address the convergence issue in AdaPID, we draw inspiration from AMSGrad \cite{reddi2018on}, which replaces the second-moment estimate with its running maximum, ensuring a non-increasing effective learning rate. We uniquely integrate this idea into AdaPID.
	
	\item[(b).] To address the stability issue in AdaPID, we derive the idea  from DiffGrad \cite{dubey2020}, which introduces a modulation (scaling) factor based on the difference between the current and previous gradient. We innovatively integrate this idea into AdaPID.
	
	\item [(c).]  We combine both the  above ideas into AdaPID, resulting in our novel Improved AdaPID AMSDiffGrad (IAdaPID-ADG) optimizer.
	
	\item [(d).] We test IAdaPID-ADG on two types of datasets. The first includes benchmark datasets, namely MNIST and CIFAR10, while the second consists of real-world cervical cancer datasets, including IARC and AnnoCerv. The performance of our optimizer is evaluated using three metrics, namely training loss, training accuracy, and testing accuracy. Across all datasets, IAdaPID-ADG achieves training losses that are multiple orders of magnitude lower than those of the competing optimizers. In addition, our optimizer achieves significantly higher training and testing accuracy with more stable curves than the existing optimizers.
	
	\item [(e).] We also conduct an ablation study on the MNIST dataset to analyze the contribution of each component of IAdaPID-ADG. The study demonstrates that every added component consistently improves the performance of our proposed optimizer.

\end{itemize}

The rest of this manuscript consists of five more sections. In Section \ref{sec:backgroud}, we review the existing literature and provide the necessary background. The proposed IAdaPID-ADG optimizer is described in Section \ref{sec:IAdaPID-ADG}. In Section \ref{sec:results}, we present the datasets and discuss the experimental results. The ablation study, examining the contribution of each component of the proposed optimizer, is provided in Section \ref{sec:ablation}. Section \ref{sec:conclusion} presents the conclusion and future directions of this work.

\section{BACKGROUND OF THE INVENTION}\label{sec:backgroud}

Deep learning optimizers are fundamentally based on gradient descent, with Stochastic Gradient Descent (SGD) serving as the core method for training neural networks \cite{robbn1951stochastic}. These networks are trained using iterative algorithms that minimize a loss function $\mathcal{L}(w)$ with respect to the parameter vector $w \in \mathbb{R}^d$. At iteration $t$, the SGD update is defined as follows:
\begin{equation}
	w_{t} = w_{t-1} - \eta g_t,
	\label{eq:sgd}
\end{equation}
where $\eta$ represents the learning rate controlling the step size and $g_t = \nabla_{w} \mathcal{L}_t(w_{t-1})$ denotes the gradient of the loss at iteration $t$. Polyak et al. \cite{polyak1964some} have shown that SGD can exhibit slow convergence and unstable updates. This is because it uses only the current gradient and does not use past information.

To address this limitation, momentum was introduced to accelerate convergence \cite{polyak1964some}. It accumulates gradients from previous iterations, enabling faster and more stable movement along relevant directions. The momentum term $m_t$ is computed as follows:
\begin{equation}
	m_t = \alpha m_{t-1} + g_t,
	\label{eq:sdg-mom}
\end{equation}
where $\alpha$ denotes the momentum coefficient that controls the contribution of past gradients. By substituting the gradient term in (\ref{eq:sgd}) with the momentum term $m_t$, the update rule becomes
\begin{equation}
	w_t = w_{t-1} - \eta m_t.
	\label{eq:sdg-mom-update}
\end{equation}
The momentum-based SGD algorithm is outlined in Algorithm~\ref{algo:sgdmomentum}.
\begin{algorithm}[!htbp]
	\caption{Momentum-Based SGD}
	\label{algo:sgdmomentum}
	\SetAlgoLined
	\textbf{Given:} Scalar $ \in \mathbb{N}_0$; $t$: iteration index; $t = 0 $ \\
	
	\textbf{Given:} Scalars $ \in \mathbb{R}$;
	\begin{itemize}
		\setlength{\itemsep}{1pt}
		\setlength{\parskip}{0pt}
		\setlength{\parsep}{0pt}
		\item $\alpha$: first moment decay rate; $\alpha  = 0.9$
		\item $\eta$: learning rate; $\eta = 0.001$
	\end{itemize}
	\textbf{Given:} Vectors $ \in \mathbb{R}^d$;
	\begin{itemize}
		\setlength{\itemsep}{1pt}
		\setlength{\parskip}{0pt}
		\setlength{\parsep}{0pt}
		\item $w_t$: parameters (weights) at iteration $t$; \\ \quad \quad $w_0$ = randomly initialized weights
		\item $g_t$: gradient
		\item $m_t$: first moment estimate; $m_0=\mathbf{0}$ 
	\end{itemize}

	\While{not converged}{
		
		1: $t = t + 1$\;\vspace{3pt}
		
		2: $g_t = \nabla_{w} \mathcal{L}_t(w_{t-1})$\; \vspace{3pt}
		
		3: $m_t = \alpha m_{t-1} + g_t$\; \vspace{3pt}
		
		4: $w_t = w_{t-1} - \eta m_t$\; 
		
	}
\end{algorithm}

Despite this improvement, momentum-based SGD still suffers from two key limitations. First, Duchi et al. \cite{duchi2011adaptive} have argued that gradients can be noisy and vary significantly across parameters, which makes a single global learning rate less effective across all parameters during training. To address this issue adaptive optimization methods were proposed. These methods dynamically adjust the step size for each parameter, resulting in smoother and less varying gradients. One such method is the Adam optimizer \cite{kingma2014adam}, which adjusts the learning rate for each parameter using the history of gradients and their squared values.

Second, Wang et al. \cite{wang2020pid} have shown that momentum-based SGD may experience an overshoot effect, where accumulated gradients cause the parameters to move beyond the optimal point, resulting in oscillations, i.e., repeated fluctuations around the optimum. To address this issue Proportional–Integral–Derivative (PID) optimization method was proposed \cite{wang2020pid}. This method is inspired by control theory and incorporates proportional, integral, and derivative terms to regulate parameter updates.

However, PID still relies on a fixed learning rate. Hence, as a result, it may not adapt effectively to noisy and varying gradient characteristics across parameters and training stages. To overcome this limitation, an adaptive variant of PID, i.e., Adaptive PID (AdaPID) \cite{weng2022adapid} was introduced. Due to the incorporation of an adaptive learning rate in PID, it enables smoother updates and reduced gradient variability.

\subsection{Adam Optimizer}

Adam is an extension of momentum-based SGD that addresses the issue of noisy and highly varying gradients across parameters by employing adaptive learning rates \cite{kingma2014adam}. The main steps of this algorithm are given below.

Initially, Adam computes an exponential moving average of the gradients, known as the \textit{first moment estimate}. Accordingly, the updated equation, which is modified from (\ref{eq:sdg-mom}), is given by
\begin{equation}
	m_t = \alpha m_{t-1} + (1-\alpha) g_t,
\end{equation}
where $\alpha$ is the exponential decay rate for the first moment estimates. This coefficient plays a role similar to the momentum coefficient defined in momentum-based SGD, as it controls the contribution of past gradients in the moving average.

Next, Adam computes an exponential moving average of the squared gradients, known as the \textit{second moment estimate}. It is computed as follows:
\begin{equation}
	v_t = \beta v_{t-1} + (1-\beta) g_t^2,
	\label{eq:secmom1}
\end{equation}
where $\beta$ is the exponential decay rate for the second moment estimates.
To correct the bias introduced during the initial steps, bias-corrected first and second moment estimates are computed as follows:
\begin{equation}
	\hat{m}_t = \frac{m_t}{1 - \alpha^t}, \quad
	\hat{v}_t = \frac{v_t}{1 - \beta^t}.
	\label{eq:biasadam}
\end{equation}

Using these bias-corrected estimates, the parameter update rule, reformulated from (\ref{eq:sdg-mom-update}), is given by
\begin{equation}
	w_{t} = w_{t-1} - \eta \frac{\hat{m}_t}{\sqrt{\hat{v}_t} + \sigma},
	\label{eq:adam-update}
\end{equation}
where $\sigma$ is a small constant added for numerical stability.
The procedure of Adam is given in Algorithm~\ref{algo:adam}.
\begin{algorithm}[!htbp]
	\caption{Adam Optimizer}
	\label{algo:adam}
	\SetAlgoLined
	\textbf{Given:} Scalar $ \in \mathbb{N}_0$; $t$: iteration index; $t = 0 $ \\
	
	\textbf{Given:} Scalars $ \in \mathbb{R}$;
	\begin{itemize}
		\setlength{\itemsep}{1pt}
		\setlength{\parskip}{0pt}
		\setlength{\parsep}{0pt}
		\item $\alpha$: first moment decay rate; $\alpha  = 0.9$
		\item $\beta$: second moment decay rate; $\beta = 0.99$
		\item $\eta$: learning rate; $\eta = 0.001$
		\item $\sigma$: numerical stability constant; $\sigma = 10^{-8}$
	\end{itemize}
	\textbf{Given:} Vectors $ \in \mathbb{R}^d$;
	\begin{itemize}
		\setlength{\itemsep}{1pt}
		\setlength{\parskip}{0pt}
		\setlength{\parsep}{0pt}
		\item $w_t$: parameters (weights) at iteration $t$; \\ \quad \quad $w_0$ = randomly initialized weights
		\item $g_t$: gradient
		\item $m_t$: first moment estimate; $m_0=\mathbf{0}$ 
		\item $v_t$: second moment estimate; $v_0=\mathbf{0}$
		\item $\hat{m}_t$: bias-corrected first moment estimate of the gradient
		\item $\hat{v}_t$: bias-corrected second moment estimate of the gradient
	\end{itemize}
	
	\While{not converged}{
		
		1: $t = t + 1$\;\vspace{3pt}
		
		2: $g_t = \nabla_{w} \mathcal{L}_t(w_{t-1})$\; \vspace{3pt}
		
		3: $m_t = \alpha m_{t-1} + (1-\alpha) g_t$\; \vspace{3pt}
		
		4: $v_t = \beta v_{t-1} + (1-\beta) g_t^2$\; \vspace{3pt}
		
		5: $\hat{m}_t = \dfrac{m_t}{1-\alpha^t}$, \quad  
		$\hat{v}_t = \dfrac{v_t}{1-\beta^t}$\;  \vspace{3pt}
		
		6: $w_t = w_{t-1} - \eta \dfrac{\hat{m}_t}{\sqrt{\hat{v}_t}+\sigma}$\;
		
	}
\end{algorithm}

\subsection{PID Optimizer}

PID-based optimization is again an extension of momentum-based SGD that addresses the oscillations problem encountered during training by introducing a derivative component, which comes from control theory \cite{wang2020pid}.


Initially, the optimizer computes an accumulated measure of past gradients in an exponentially decayed manner, referred to as the \textit{integral term}, which is derived from (\ref{eq:sdg-mom}). It is given by
\begin{equation}
	I_t = \gamma I_{t-1} - \eta g_t,
	\label{eq:integ}
\end{equation}
where $\gamma$ is a scaling factor.

Next, the optimizer computes the rate of change of gradients between consecutive iterations, referred to as the \textit{derivative term}, defined as an exponential moving average of gradient differences. It is computed as follows:
\begin{equation}
	D_t = \gamma D_{t-1} + (1-\gamma)(g_t - g_{t-1}).
	\label{eq:deriv}
\end{equation}

Using these terms, the parameter update rule, in place of (\ref{eq:sdg-mom-update}), is given by
\begin{equation}
	w_t = w_{t-1} + I_t + K_d D_t,
	\label{eq:pid-update}
\end{equation}
where $K_d$ is the derivative gain coefficient. The PID optimization method is given in Algorithm~\ref{algo:pid}.
\begin{algorithm}[!htbp]
	\caption{PID Optimizer}
	\label{algo:pid}
	\SetAlgoLined
	\textbf{Given:} Scalar $ \in \mathbb{N}_0$; $t$: iteration index; $t = 0 $ \\
	
	\textbf{Given:} Scalars $ \in \mathbb{R}$;
	\begin{itemize}
		\setlength{\itemsep}{1pt}
		\setlength{\parskip}{0pt}
		\setlength{\parsep}{0pt}
		\item $\gamma$: scaling factor for integral and derivative terms; $\gamma  = 0.9$
		\item $\eta$: learning rate; $\eta = 0.001$
		\item $K_d$: derivative gain coefficient; $K_d = 1$
	\end{itemize}
	\textbf{Given:} Vectors $ \in \mathbb{R}^d$;
	\begin{itemize}
		\setlength{\itemsep}{1pt}
		\setlength{\parskip}{0pt}
		\setlength{\parsep}{0pt}
		\item $w_t$: parameters (weights) at iteration $t$; \\ \quad \quad $w_0$ = randomly initialized weights
		\item $g_t$: gradient
		\item $I_t$: integral term; $I_0=\mathbf{0}$ 
		\item $D_t$: derivative term; $D_0=\mathbf{0}$
	\end{itemize}

	\While{not converged}{
		
		1: $t = t + 1$\;\vspace{3pt}
		
		2: $g_t = \nabla_{w} \mathcal{L}_t(w_{t-1})$\; \vspace{3pt}
		
		3: $I_t = \gamma I_{t-1} - \eta g_t$\; \vspace{3pt}
		
		4: $\Delta g_t = g_t - g_{t-1}$\; \vspace{3pt}
		
		5: $D_t = \gamma D_{t-1} + (1-\gamma)\Delta g_t$\; \vspace{3pt}
		
		6: $w_t = w_{t-1} + I_t + K_d D_t$\; 
		
	}
\end{algorithm}

\subsection{Adaptive PID (AdaPID) Optimizer }

Adaptive PID (AdaPID) combines the Adam and the PID optimization techniques. As a result, noisy and highly varying gradients are handled through adaptive learning rates, while oscillations are mitigated by the inclusion of the derivative component \cite{weng2022adapid}.

The base algorithm here is PID and the components of Adam are added to it. Accordingly, the first step involves the computation of the \textit{integral term}, which accumulates past gradients in an exponentially decayed manner. In this formulation, the learning rate is not included in this term, unlike its inclusion in the PID formulation. This is because the learning rate affects both the integral and derivative components and is therefore incorporated later in the final parameter update rule. The integral term, which is updated from (\ref{eq:integ}), is defined as
\begin{equation}
	I_t = \gamma I_{t-1} + g_t.
\end{equation}
Next, the derivative term, which captures the change in gradients between consecutive iterations, is computed as an exponential moving average of gradient differences, is retained as in (\ref{eq:deriv}), and is defined as
\begin{equation}
	D_t = \gamma D_{t-1} + (1-\gamma)(g_t - g_{t-1}).
\end{equation}

Now, the aspect of adaptiveness is incorporated. Specifically, Adam computes the exponential moving average of the squared gradients, as defined in (\ref{eq:secmom1}), given by
\begin{equation}
	v_t = \beta v_{t-1} + (1-\beta) g_t^2,
	\label{eq:secmom2_1}
\end{equation}
where $\beta$ is a scaling factor. AdaPID technique adds another term, which captures the exponential moving average of the squared gradient differences. This is given as follows:
\begin{equation}
	d_t = \beta d_{t-1} + (1-\beta) (g_t - g_{t-1})^2.
	\label{eq:secmom2_2}
\end{equation}
The bias-corrected estimates of these terms are given by
\begin{equation}
	\hat{v}_t = \frac{v_t}{1-\beta^t}, \quad
	\hat{d}_t = \frac{d_t}{1-\beta^t}.
	\label{eq:bias_ada}
\end{equation}

Using these quantities, the parameter update rule, reformulated from (\ref{eq:adam-update}) and (\ref{eq:pid-update}), is given by
\begin{equation}
	w_{t} = w_{t-1} - \eta \left(
	\frac{K_i I_t}{\sqrt{\hat{v}_t} + \sigma} +
	\frac{K_d D_t}{\sqrt{\hat{d}_t} + \sigma}
	\right),
	\label{eq:adapid}
\end{equation}
where $K_i$ and $K_d$ integral and derivative gain coefficients, respectively.
\begin{algorithm}[!htbp]
	\caption{Adaptive PID (AdaPID) Optimizer}
	\label{algo:adapid}
	\SetAlgoLined
	\textbf{Given:} Scalar $ \in \mathbb{N}_0$; $t$: iteration index; $t = 0 $ \\
	
	\textbf{Given:} Scalars $ \in \mathbb{R}$;
	\begin{itemize}
		\setlength{\itemsep}{1pt}
		\setlength{\parskip}{0pt}
		\setlength{\parsep}{0pt}
		\item $\gamma$: scaling factor for integral and derivative terms; $\gamma  = 0.9$
		\item $\beta$: scaling factor for squared of gradient and squared and gradient difference terms; $\beta = 0.99$
		\item $\eta$: learning rate; $\eta = 0.001$
		\item $K_i$: integral gain coefficient; $K_i = 0.5$
		\item $K_d$: derivative gain coefficient; $K_d = 1$
		\item $\sigma$: numerical stability constant; $\sigma = 10^{-8}$
	\end{itemize}
	\textbf{Given:} Vectors $ \in \mathbb{R}^d$;
	\begin{itemize}
		\setlength{\itemsep}{1pt}
		\setlength{\parskip}{0pt}
		\setlength{\parsep}{0pt}
		\item $w_t$: parameters (weights) at iteration $t$; \\ \quad \quad $w_0$ = randomly initialized weights
		\item $g_t$: gradient
		\item $I_t$: integral term; $I_0=\mathbf{0}$ 
		\item $D_t$: derivative term; $D_0=\mathbf{0}$
		\item $v_t$: exponential moving average of gradient square; $v_0=\mathbf{0}$ 
		\item $d_t$: exponential moving average of gradient difference square; $d_0=\mathbf{0}$
		\item $\hat{v}_t$: bias-corrected $v_t$
		\item $\hat{d}_t$: bias-corrected $d_t$
	\end{itemize}

	\While{not converged}{
		
		1: $t = t + 1$\;\vspace{3pt}
		
		2: $g_t = \nabla_{w} \mathcal{L}_t(w_{t-1})$\; \vspace{3pt}
		
		3: $I_t = \gamma I_{t-1} + g_t$\; \vspace{3pt}
		
		4: $\Delta g_t = g_t - g_{t-1}$\; \vspace{3pt}
		
		5: $D_t = \gamma D_{t-1} + (1-\gamma)\Delta g_t$\; \vspace{3pt}
		
		6: $v_t = \beta v_{t-1} + (1-\beta) g_t^2$\; \vspace{3pt}
		
		7: $d_t = \beta d_{t-1} + (1-\beta) (\Delta g_t)^2$\; \vspace{3pt}
		
		8: $\hat{v}_t = \dfrac{v_t}{1-\beta^t}$, \quad
		$\hat{d}_t = \dfrac{d_t}{1-\beta^t}$\; \vspace{3pt}
		
		9: $w_t = w_{t-1} - \eta \left(
		\frac{K_i I_t}{\sqrt{\hat{v}_t} + \sigma}
		+
		\frac{K_d D_t}{\sqrt{\hat{d}_t} + \sigma}
		\right)$\;
		
	}
\end{algorithm}

\section{Proposed IAdaPID-ADG Optimizer}\label{sec:IAdaPID-ADG}
In this section, we present the technical description of the proposed optimizer. Although the AdaPID optimizer performs well, it still carries two drawbacks inherited from Adam. \textit{First}, it may fail to achieve convergence \cite{reddi2018on}. \textit{Second}, it may lead to instability during training \cite{dubey2020}. To address the convergence issue, we draw inspiration from the idea proposed in AMSGrad \cite{reddi2018on} and integrate it with AdaPID. Similarly, to address the stability issue, we draw intuition from the concept introduced in DiffGrad \cite{dubey2020} and integrate it with AdaPID. Finally, to address both issues simultaneously, we combine these ideas with AdaPID and term the resulting optimizer as Improved AdaPID AMSDiffGrad (IAdaPID-ADG). The details for each of these optimizers are discussed in the following sections.

\subsection{AdaPID + AMSGrad Optimizer}
As mentioned earlier, Adam suffers from the drawback of failing to guarantee convergence, which motivated the proposal of AMSGrad \cite{reddi2018on}. The core idea behind AMSGrad is to replace the exponential moving average of squared gradients and its bias corrected form given in (\ref{eq:secmom1}) and (\ref{eq:biasadam}) in Adam, with a maximum operation. This is defined as follows \cite{reddi2018on}: 
\begin{equation}
	v_t^{\text{max}} = \max(v_{t-1}^{max}, v_t).
	\label{eq:max1}
\end{equation}
The corresponding bias corrected form is now given by
\begin{equation}
	\hat{v}_t^{\max} = \dfrac{v_t^{\max}}{1-\beta^t}.
	\label{eq:bias_max_1}
\end{equation}
We apply the same max operation to AdaPID as well, i.e., instead of using (\ref{eq:secmom2_1}) and (\ref{eq:bias_ada}), we use (\ref{eq:max1}) and (\ref{eq:bias_max_1}).

It is worth noting that AdaPID, in addition to $v_t$, also maintains $d_t$, which represents the exponential moving average of squared gradient differences. Since the algorithm is symmetric with respect to the squared gradient and the squared gradient difference, we apply the same max rule to $d_t$, i.e., instead of directly using (15) and (14) in AdaPID, we use
\begin{equation}
	d_t^{\text{max}} = \max(d_{t-1}^{max}, d_t).
\end{equation}
The corresponding bias corrected form is now given by
\begin{equation}
	\hat{d}_t^{\max} = \dfrac{d_t^{\max}}{1-\beta^t}.
\end{equation}
The final update rule, analogous to (\ref{eq:adapid}) in AdaPID, is given by
\begin{equation}
	w_{t} = w_{t-1} - \eta \left(
	\frac{K_i I_t}{\sqrt{\hat{v}_t^{\text{max}}} + \sigma}
	+
	\frac{K_d D_t}{\sqrt{\hat{d}_t^{\text{max}}} + \sigma}
	\right).
	\label{eq:adaams}
\end{equation}

\subsection{AdaPID + DiffGrad Optimizer}
As discussed earlier, the second limitation of the Adam optimizer is related to instability during training. To address this limitation, the DiffGrad optimizer was proposed. The key idea behind DiffGrad is that it introduce an new modulation factor which measures the difference between gradients of consecutive iterations. To this end, the gradient difference is first computed as
\begin{equation}
	\Delta g_t = g_t - g_{t-1}.
\end{equation}
Based on this difference, the modulation factor is defined as \cite{dubey2020}
\begin{equation}
\mu_{t} = \dfrac{1}{1 + e^{-\left|\Delta g_{t}\right|}}.
\end{equation}
Finally, this factor $\mu_t$ is multiplied with the learning rate $\eta$ in (\ref{eq:adam-update}). We apply the same idea to AdaPID, i.e., the update rule, instead of (\ref{eq:adapid}), is given by
\begin{equation}
	w_t = w_{t-1} - \eta \mu_t \left(
	\frac{K_i I_t}{\sqrt{\hat{v}_t} + \sigma}
	+
	\frac{K_d D_t}{\sqrt{\hat{d}_t} + \sigma}
	\right).
	\label{eq:adadiff}
\end{equation}

\subsection{AdaPID + AMSGrad + DiffGrad (IAdaPID-ADG) Optimizer}
In the proposed optimizer, we integrate the principles of AMSGrad and DiffGrad with AdaPID to achieve improved convergence and stability, and term the resulting optimizer as IAdaPID-ADG. The procedure of our IAdaPID-ADG optimizer is given in Algorithm~\ref{algo:IAdaPID-ADG}.
\begin{algorithm}[!htbp]
	\caption{IAdaPID-ADG Optimizer}
	\label{algo:IAdaPID-ADG}
	\SetAlgoLined
\textbf{Given:} Scalar $ \in \mathbb{N}_0$; $t$: iteration index; $t = 0 $ \\

\textbf{Given:} Scalars $ \in \mathbb{R}$;
\begin{itemize}
	\setlength{\itemsep}{1pt}
	\setlength{\parskip}{0pt}
	\setlength{\parsep}{0pt}
	\item $\gamma$: scaling factor for integral and derivative terms; $\gamma  = 0.9$
	\item $\beta$: scaling factor for squared of gradient and squared and gradient difference terms; $\beta = 0.99$
	\item $\eta$: learning rate; $\eta = 0.001$
	\item $K_i$: integral gain coefficient; $K_i = 0.5$
	\item $K_d$: derivative gain coefficient; $K_d = 1$
	\item $\sigma$: numerical stability constant; $\sigma = 10^{-8}$
\end{itemize}
\textbf{Given:} Vectors $ \in \mathbb{R}^d$;
\begin{itemize}
	\setlength{\itemsep}{1pt}
	\setlength{\parskip}{0pt}
	\setlength{\parsep}{0pt}
	\item $w_t$: parameters (weights) at iteration $t$; \\ \quad \quad $w_0$ = randomly initialized weights
	\item $g_t$: gradient
	\item $\mu_t$: DiffGrad modulation factor
	\item $I_t$: integral term; $I_0=\mathbf{0}$ 
	\item $D_t$: derivative term; $D_0=\mathbf{0}$
	\item $v_t$: exponential moving average of gradient square; $v_0=\mathbf{0}$ 
	\item $d_t$: exponential moving average of gradient difference square; $d_0=\mathbf{0}$
	\item $\hat{v}_t$: bias-corrected $v_t$
	\item $\hat{d}_t$: bias-corrected $d_t$
	\item $\hat{v}_t^{\max}$: maximum of $\hat{v_t}$; $\hat{v}_0^{\max}=\mathbf{0}$
	\item $\hat{d}_t^{\max}$: maximum of $\hat{d_t}$; $\hat{d}_0^{\max}=\mathbf{0}$
\end{itemize}

	\While{not converged}{
		
		1: $t = t + 1$\;\vspace{3pt}
		
		2: $g_t = \nabla_{w} \mathcal{L}_t(w_{t-1})$\; \vspace{3pt}
		
		3: $\Delta g_t = g_t - g_{t-1}$\; \vspace{3pt}
		
		4: $\mu_t = \dfrac{1}{1 + e^{-|\Delta g_t|}}$\; \vspace{3pt}
		
		5: $I_t = \gamma I_{t-1} + g_t$\; \vspace{3pt}
		
		6: $D_t = \gamma D_{t-1} + (1-\gamma)\Delta g_t$\; \vspace{3pt}
		
		7: $v_t = \beta v_{t-1} + (1-\beta) g_t^2$\; \vspace{3pt}
		
		8: $d_t = \beta d_{t-1} + (1-\beta) (\Delta g_t)^2$\; \vspace{3pt}

		9: $v_t^{\max} = \max(v_{t-1}^{\max}, v_t),\quad
		d_t^{\max} = \max(d_{t-1}^{\max}, d_t)$\; \vspace{3pt}
		
		10: $\hat{v}_t^{\max} = \dfrac{v_t^{\max}}{1-\beta^t}, \quad
		\hat{d}_t^{\max} = \dfrac{d_t^{\max}}{1-\beta^t}$\; \vspace{3pt}
		
		11: $w_t = w_{t-1} - \eta \,\mu_t \left(
		\frac{K_i I_t}{\sqrt{\hat{v}_t^{\max}} + \sigma}
		+
		\frac{K_d D_t}{\sqrt{\hat{d}_t^{\max}} + \sigma}
		\right)$\;
		
	}
	
	\vspace{6pt}
	\noindent\rule{\linewidth}{0.4pt}\\[2pt]
	{\footnotesize
		$^{\#}$\textbf{Proposed modifications over AdaPID:}\\
		\textbf{Step 4:} $\mu_t$ is the DiffGrad modulation factor, computed as the sigmoid of the absolute gradient difference.\\
		\textbf{Step 9:} $v_t^{\max}$ and $d_t^{\max}$ are element-wise running maxima of $v_t$ and $d_t$ respectively, inspired by AMSGrad, ensuring non-increasing effective learning rates.\\
		\textbf{Step 10:} Bias-corrected maxima $\hat{v}_t^{\max}$ and $\hat{d}_t^{\max}$ replace standard bias-corrected estimates in the denominator.\\
		\textbf{Step 11:} The weight update applies $\mu_t$ as a modulation factor and uses $\sqrt{\hat{v}_t^{\max}}$ and $\sqrt{\hat{d}_t^{\max}}$ in the denominators.
	}
\end{algorithm}

\section{EXPERIMENTAL RESULTS}\label{sec:results}

This section is organized into two subsections. Section \ref{subsec: benchmark} presents the performance of the proposed optimizer on the benchmark datasets MNIST \cite{lecun1998gradient} and CIFAR10 \cite{krizhevsky2009learning}, whereas the results on the real-world medical datasets IARC \cite{IARC2024} and AnnoCerv \cite{Minciuna2023} are discussed in Section \ref{subsec: real-life}. We present the performance metrics of the best-performing models obtained across multiple runs using different hyperparameter configurations, and these hyperparameter configurations for all optimizers are provided in Table~\ref{tab:optimizer_params}.

\subsection{Results on Benchmark Datasets}\label{subsec: benchmark}
As mentioned above, we evaluate our optimizer on the MNIST \cite{lecun1998gradient} and CIFAR-10 \cite{krizhevsky2009learning} datasets, which are widely used benchmarks in the machine learning and computer vision communities for assessing optimization performance. On both datasets, we compare the performance of multiple optimizers, including AMSGrad \cite{reddi2018on}, diffGrad \cite{dubey2020}, and AdaPID \cite{weng2022adapid}, against our proposed IAdaPID-ADG. 

\renewcommand{\arraystretch}{1.3}
\begin{table*}[!htbp]
	\caption{Hyperparameter settings of the different optimizers.}
	\centering
	\resizebox{0.9\textwidth}{!}{   
		\begin{tabular}{c|l|l|l|l|l|l|l}
			\hline
			\multirow{2}{*}{Optimizer} & \multicolumn{7}{c}{Parameters} \\ 
			\cline{2-8}
			& $\eta$ & $\alpha/\gamma$ & $\beta$ & AMSGrad & $K_i$ & $K_d$ & $\sigma$ \\
			\hline \hline
			AMSGrad \cite{reddi2018on} 
			& $\{1e\!-\!4, 5e\!-\!4, 1e\!-\!3\}$ 
			& $\{0.9, 0.99\}$
			& $\{0.99, 0.999\}$
			& True 
			& \textbf{--} 
			& \textbf{--} 
			& $10^{-8}$ \\
			
			DiffGrad \cite{dubey2020} 
			& $\{1e\!-\!4, 5e\!-\!4, 1e\!-\!3\}$ 
			& $\{0.9, 0.99\}$
			& $\{0.99, 0.999\}$
			& \textbf{--} 
			& \textbf{--} 
			& \textbf{--} 
			& $10^{-8}$ \\
			
			AdaPID \cite{weng2022adapid} 
			& $\{1e\!-\!4, 5e\!-\!4, 1e\!-\!3\}$ 
			& $\{0.9, 0.99\}$
			& $\{0.99, 0.999\}$
			& \textbf{--} 
		    & $\{0.5, 1, 5\}$
		    & $\{1, 10, 20\}$
			& $10^{-8}$ \\
			
			IAdaPID-ADG
			& $\{1e\!-\!4, 5e\!-\!4, 1e\!-\!3\}$ 
			& $\{0.9, 0.99\}$
			& $\{0.99, 0.999\}$
			& True
			& $\{0.5, 1, 5\}$
			& $\{1, 10, 20\}$
			& $10^{-8}$ \\
			\hline
	\end{tabular}}
	\label{tab:optimizer_params}
	\begin{tablenotes}
		\footnotesize
		\item $\eta$ denotes the learning rate, $\alpha$ is the exponential decay rates for the first estimates, $\gamma$ is the scaling factor for integral and derivative terms, $\beta$ is the exponential decay rates for the second moment estimates for Adam and a scaling factor for gradient square and gradient difference square terms, $K_i$ and $K_d$ represent the integral and derivative gain coefficients, and $\sigma$ is a small constant used for numerical stability.
	\end{tablenotes}
\end{table*}

\subsubsection{MNIST Dataset} \label{subsubsec:mnist}
The MNIST dataset \cite{lecun1998gradient}, a subset of the larger NIST dataset, consists of 70,000 grayscale images of handwritten digits ranging from 0 to 9, with $60,000$ images used for training and $10,000$ for testing. Each digit is size-normalized and centered within a fixed-size image of 28 × 28 pixels.

For underlying model, we follow the settings as used in the literature \cite{kingma2014adam, wang2020pid}. It is a Multi-Layer Neural Network (MLNN) comprising two fully connected hidden layers with $1000$ neurons each. ReLU activation functions are employed to introduce non-linearity, and a dropout rate of $0.3$ is applied for regularization. The model is trained with a batch size of $128$ for $20$ epochs.



\begin{figure*}[!htbp]
	\centering
	\subfloat[]{%
		\includegraphics[width=0.33\textwidth]{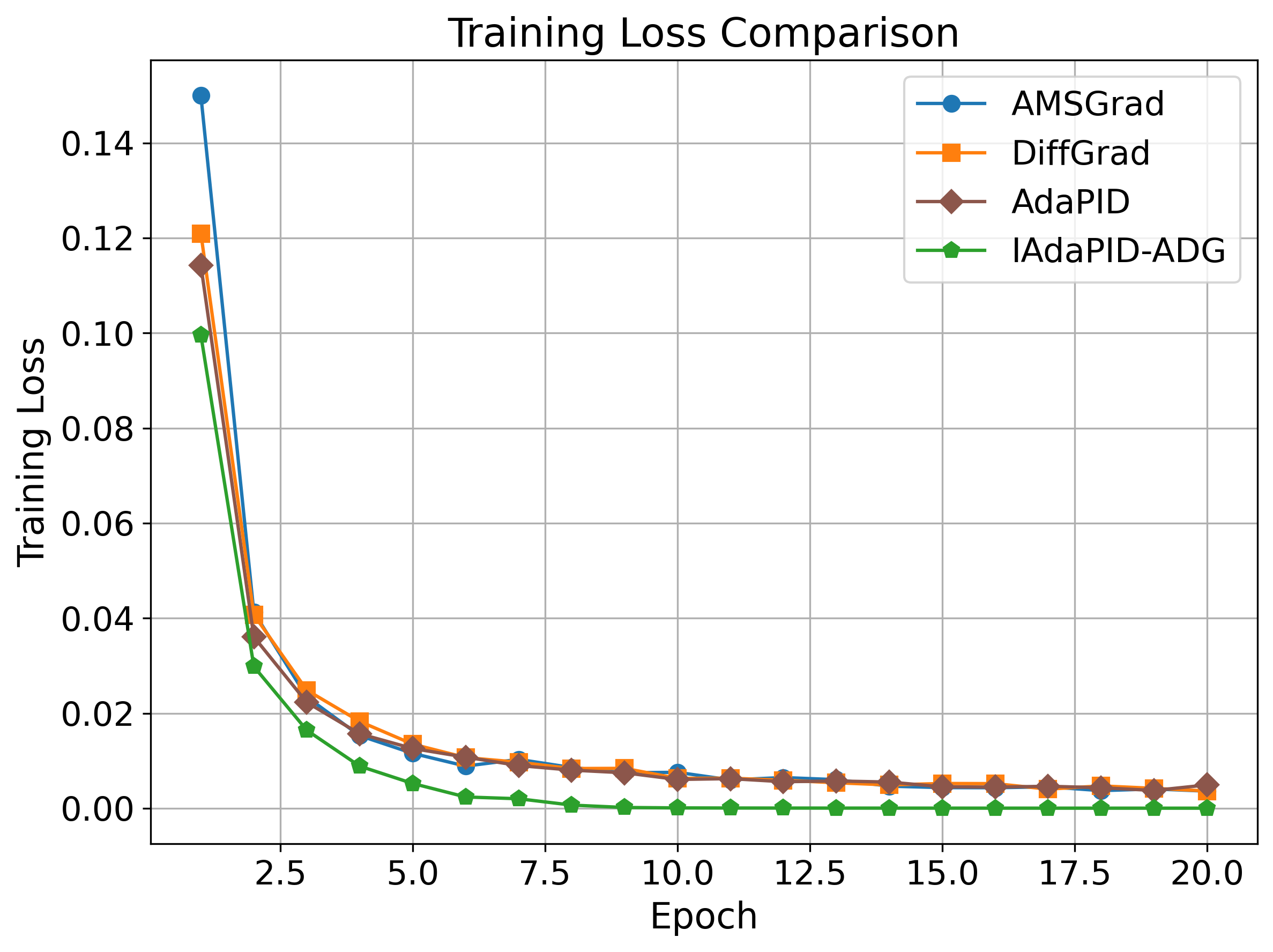}%
	}
	\hfill
	\subfloat[]{%
		\includegraphics[width=0.33\textwidth]{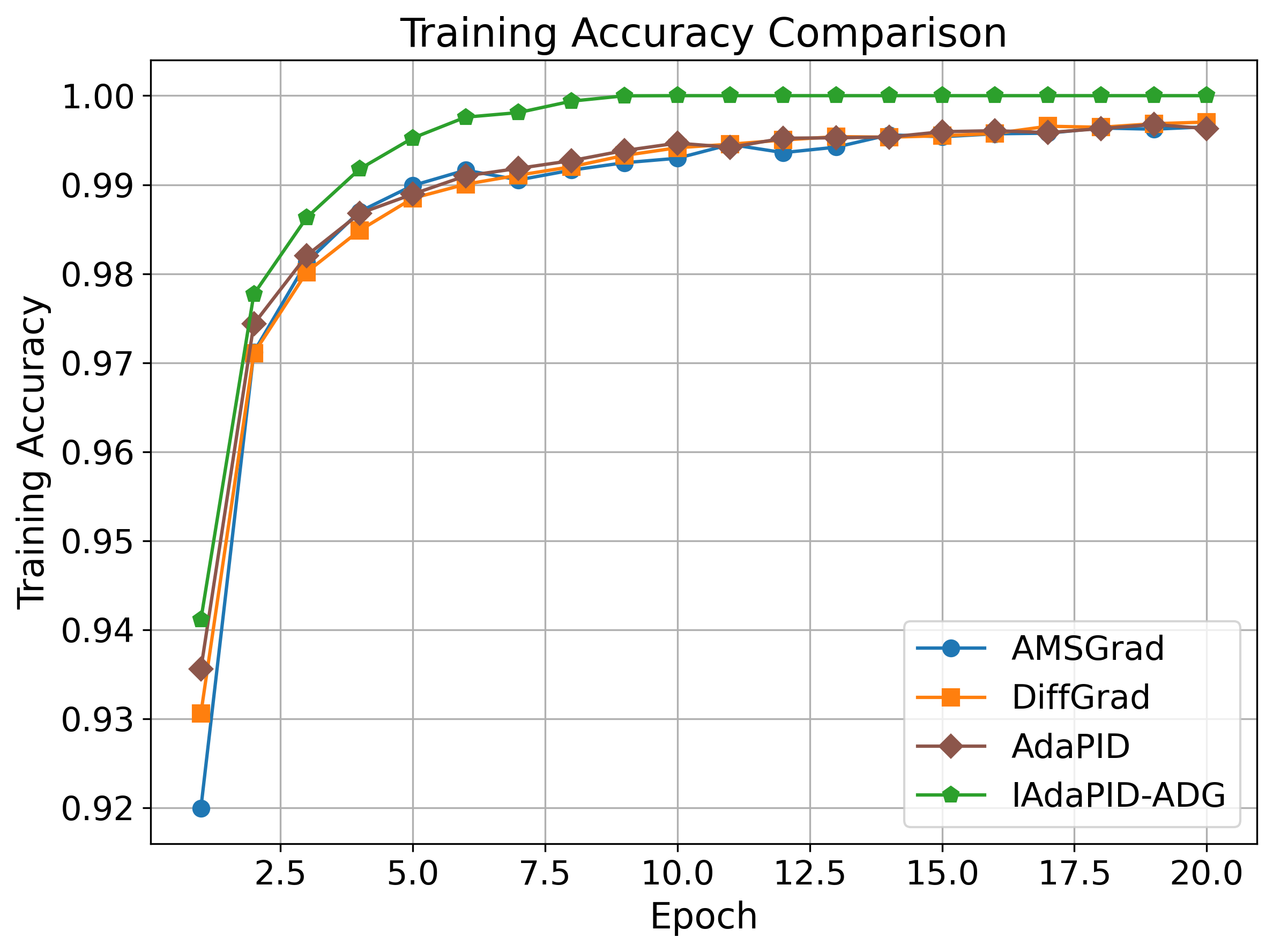}%
	}
	\hfill
	\subfloat[]{%
		\includegraphics[width=0.33\textwidth]{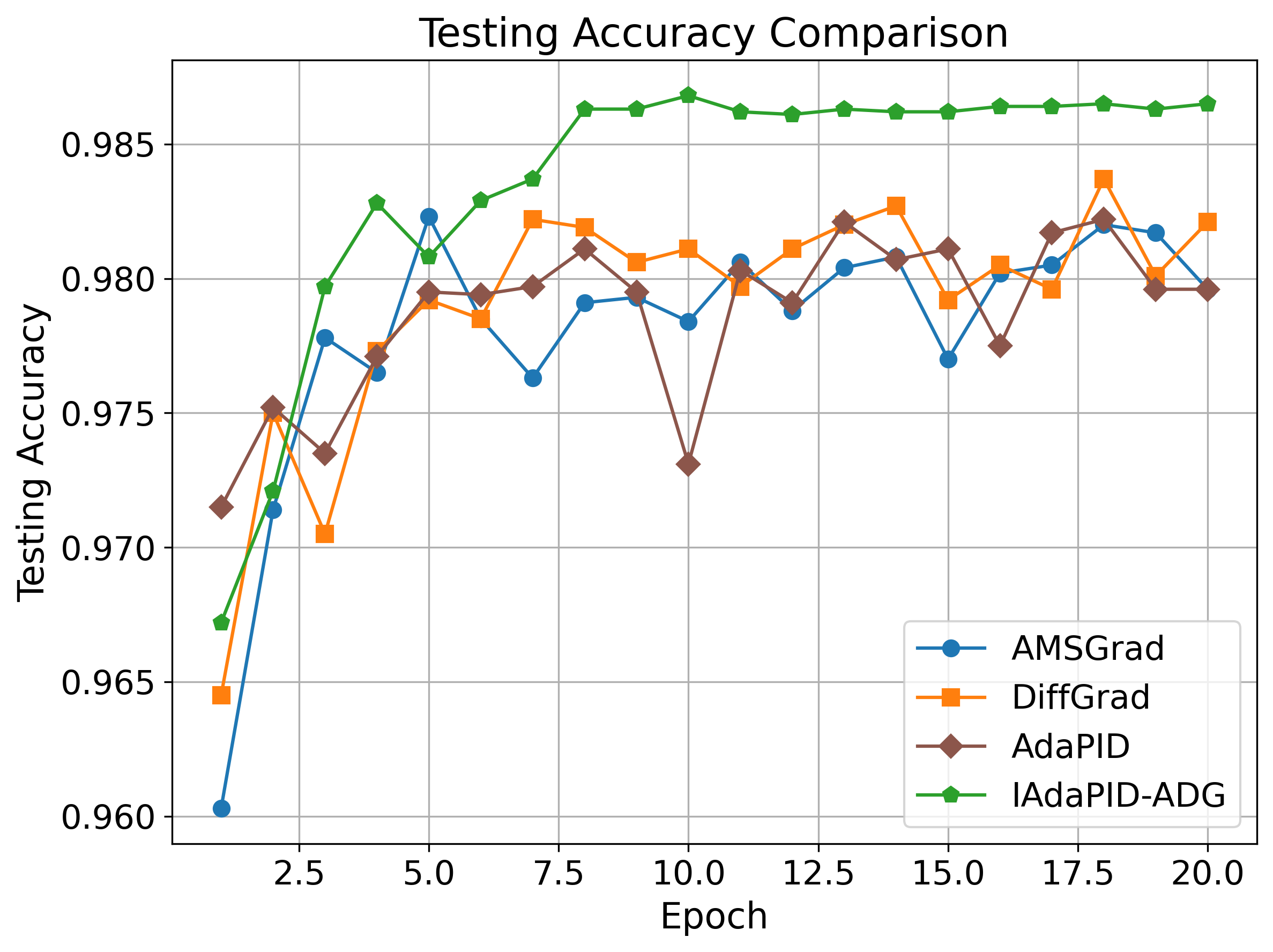}%
	}
	\caption{Comparison of AMSGrad, DiffGrad, AdaPID and IAdaPID-ADG optimizers on the MNIST dataset; (a) training loss, (b) training accuracy, and (c) testing accuracy.}
	\label{Fig: MNIST}
\end{figure*}

Table~\ref{Tab: MNIST} summarizes the performance of different optimizers, with the corresponding learning curves illustrated in Fig.~\ref{Fig: MNIST}. As shown in the table, the proposed IAdaPID-ADG optimizer delivers the best overall performance compared to the other three optimizers. It achieves a significantly lower training loss of $0.00002$, and the highest training and testing accuracies of $100\%$ and $98.68\%$, respectively. The other optimizers, namely AMSGrad, DiffGrad, and AdaPID, show comparatively inferior performance in terms of both loss and accuracy. From the Fig.~\ref{Fig: MNIST}, the curve of the proposed optimizer lies below for training loss and above for training and testing accuracies, respectively.

\renewcommand{\arraystretch}{1.3}
\begin{table}[!htbp]
	\caption{Comparison of AMSGrad, DiffGrad, AdaPID and IAdaPID-ADG optimizers on the MNIST dataset.}
	\centering
	\scriptsize
	\resizebox{0.95\columnwidth}{!}{   
		\begin{tabular}{l|l|l|l}
			\hline 
			\multirow{2}{*}{Optimizer} & \multicolumn{2}{c|}{Training} & Testing \\ \cline{2-3}
			& Loss & Accuracy (\%)& Accuracy (\%)\\
			\hline
			AMSGrad         &  0.004& 99.65 & 98.15\\
			DiffGrad    & 0.004& 99.70 &98.22  \\ 
			AdaPID    &  0.004 & 99.67 & 98.11 \\ 
			\textbf{IAdaPID-ADG} & \textbf{0.00002} &  \textbf{100}& \textbf{98.68} \\ 
			\hline
		\end{tabular}
	}
	\label{Tab: MNIST}
\end{table}






\subsubsection{CIFAR10 Dataset}\label{subsubsec:cifar}
The CIFAR10 \cite{krizhevsky2009learning} dataset consists of $60,000$ color images, divided into 50,000 training and $10,000$ testing samples. The dataset includes 10 classes, namely airplane, automobile, bird, cat, deer, dog, frog, horse, ship, and truck, with $6,000$ images per class. Each image has a resolution of $32 \times 32$ pixels.

For the underlying model, we adopt the configuration reported in the literature \cite{kingma2014adam, weng2022adapid}. The model is a Convolutional Neural Network (CNN) consisting of five convolutional layers equipped with ReLU activation and max pooling, followed by a fully connected layer, in line with \cite{weng2022adapid}. Training is performed using a batch size of $128$ for $100$ epochs.

\begin{figure*}[!htbp]
	\centering
	\subfloat[]{%
		\includegraphics[width=0.33\textwidth]{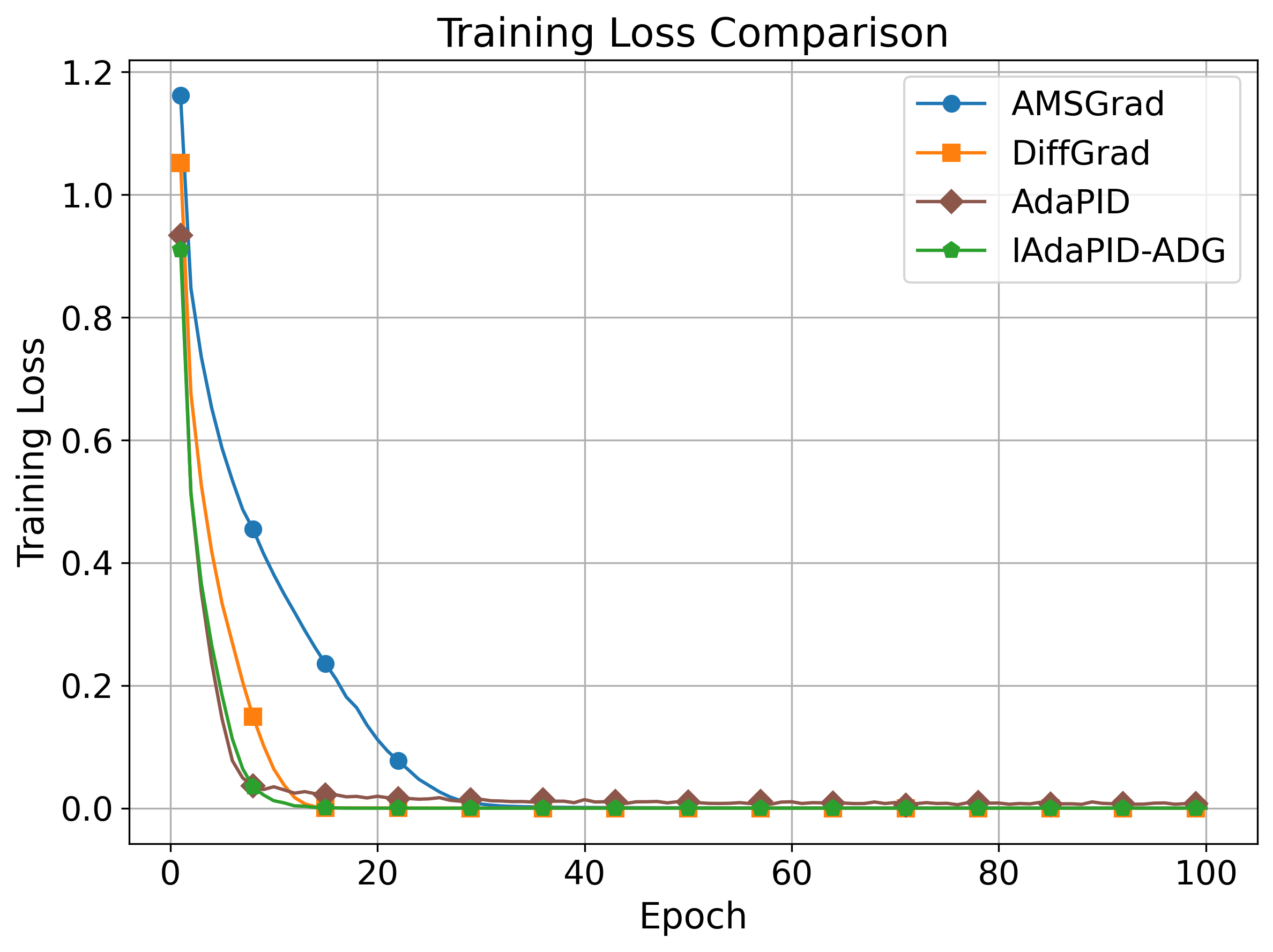}%
	}
	\hfill
	\subfloat[]{%
		\includegraphics[width=0.33\textwidth]{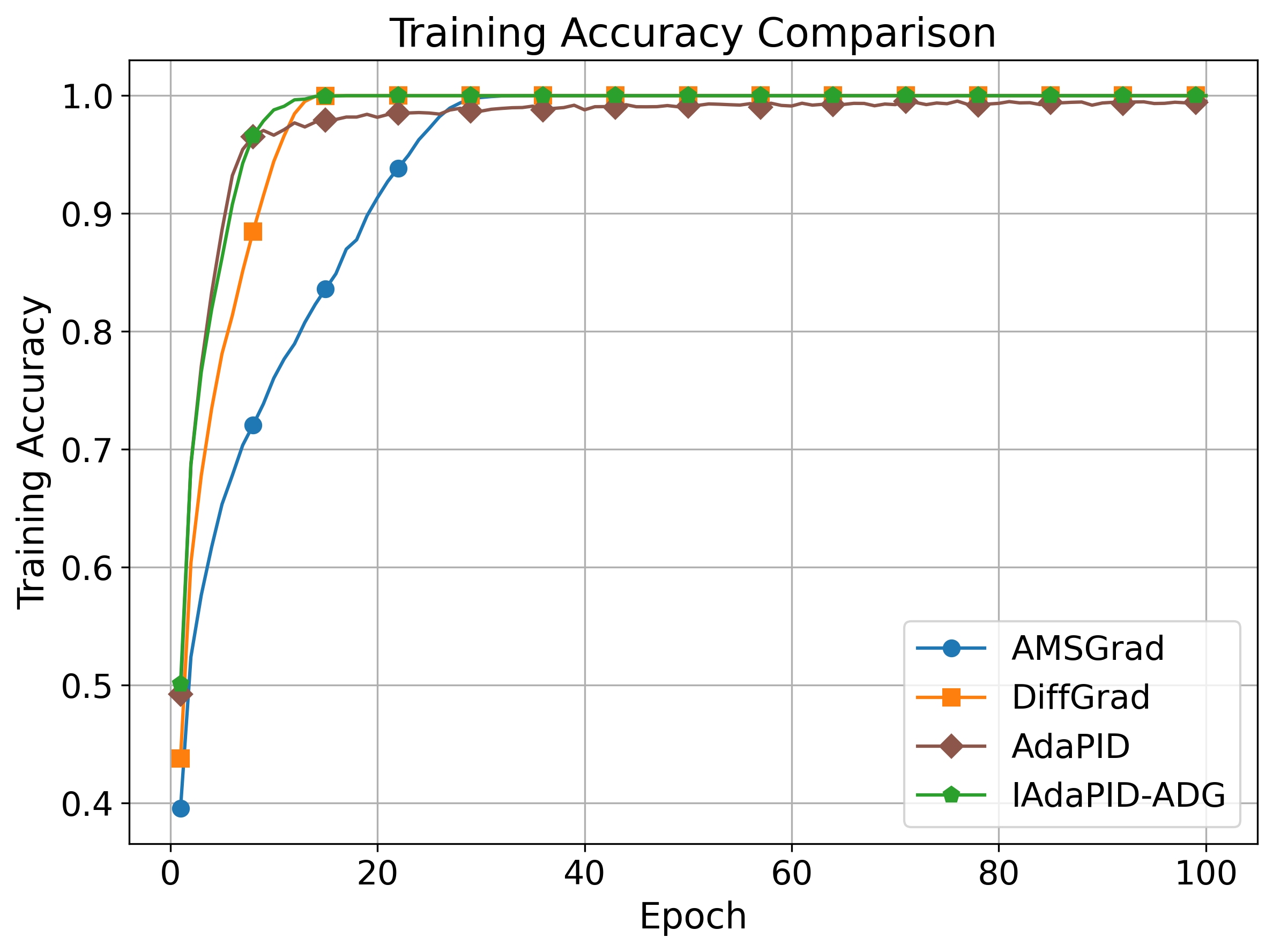}%
	}
	\hfill
	\subfloat[]{%
		\includegraphics[width=0.33\textwidth]{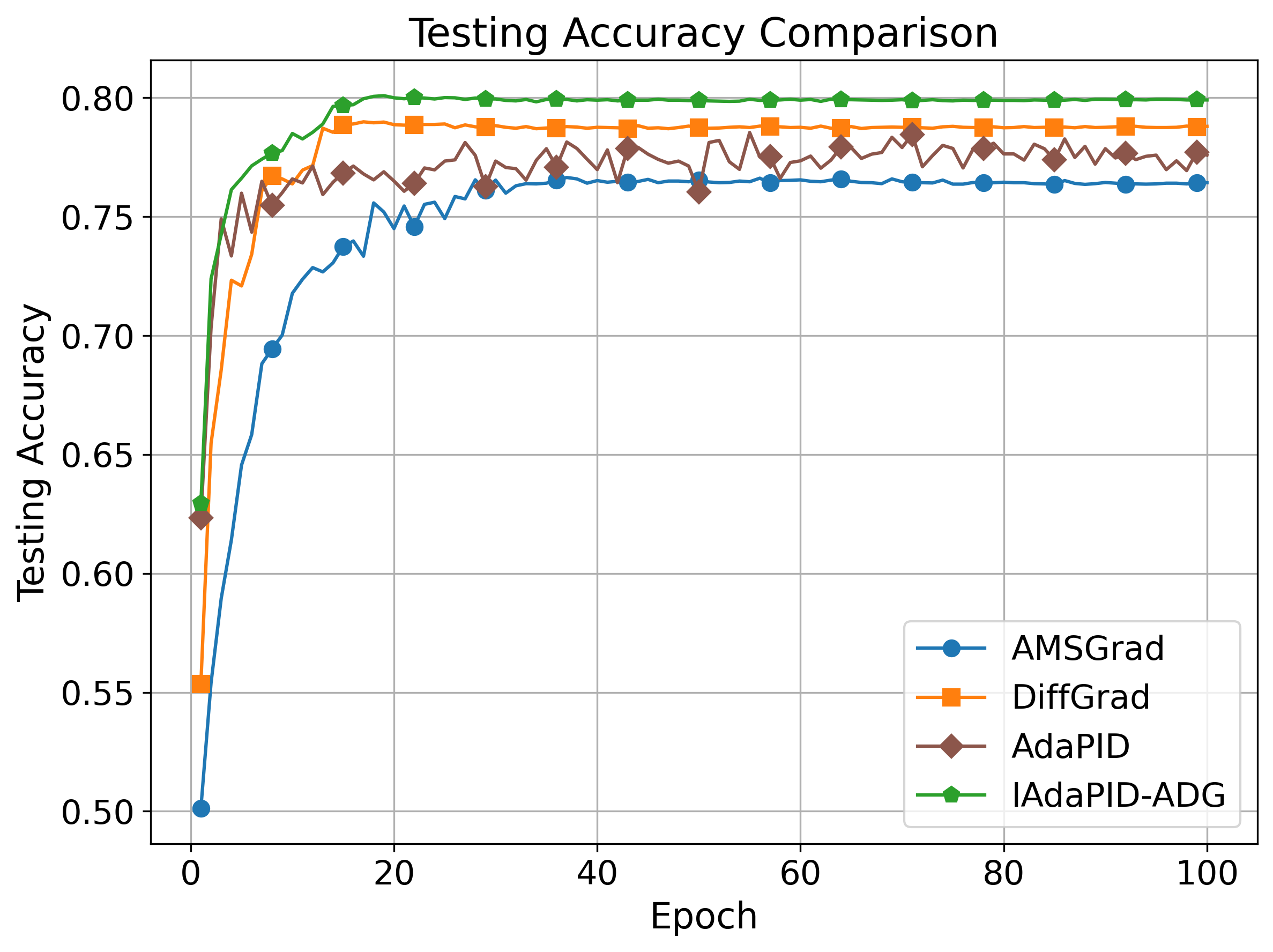}%
	}
	\caption{Comparison of AMSGrad, DiffGrad, AdaPID and IAdaPID-ADG optimizers on the CIFAR10 dataset; (a) training loss, (b) training accuracy, and (c) testing accuracy.}
	\label{Fig: CIFAR_1}
\end{figure*}

Table~\ref{Tab: CIFAR_1} shows the performance comparison of different optimizers on the CIFAR-10 dataset, along with the learning curves in Fig.~\ref{Fig: CIFAR_1}. It can be seen that the proposed IAdaPID-ADG optimizer performs better overall than the other methods. Specifically, it achieves the trough training loss of $0.000003$, while also reaching a peak training  and testing accuracy of $100\%$ and $80.09\%$, respectively. In contrast, the other optimizers, i.e., AMSGrad, DiffGrad, and AdaPID, yield comparatively higher losses and lower accuracies. From Fig.~\ref{Fig: CIFAR_1}, it is also evident that the proposed optimizer maintains a lower curve for training loss and higher curves for both training and testing accuracies. 

These results on both the benchmark datasets demonstrate the superior convergence capability and improved generalization performance of the proposed optimizer.

\renewcommand{\arraystretch}{1.3}

\begin{table}[!htbp]
	\caption{Comparison of AMSGrad, DiffGrad, AdaPID and IAdaPID-ADG optimizers on the CIFAR10 dataset.}
	\centering
		\resizebox{0.95\columnwidth}{!}{ 
\begin{tabular}{l|l|l|l}
	\hline 
	\multirow{2}{*}{Optimizer} & \multicolumn{2}{c|}{Training} & Testing \\ \cline{2-3}
	& Loss & Accuracy (\%)& Accuracy (\%)\\
	\hline
	AMSGrad     &0.0001&100&76.65\\
	DiffGrad    &0.00002&100& 78.99\\ 
	AdaPID   &0.006&99.51& 78.54\\ 
	\textbf{IAdaPID-ADG} & \textbf{0.000003} &  \textbf{100}& \textbf{80.09} \\ 
	\hline
\end{tabular}
}
	\label{Tab: CIFAR_1}
\end{table}

\subsection{Results on Real-World Medical Datasets}\label{subsec: real-life}
The results for the IARC \cite{IARC2024} and the AnnoCerv \cite{Minciuna2023} datasets are presented in the below subsections. 

In one of our recent works \cite{saini2025}, we employed three ResNet models, namely ResNet50, ResNet101, and ResNet152, for feature extraction and a Support Vector Machine (SVM) for classification. However, since our optimizer cannot be applied to SVM, we instead adopt a two-layer feed-forward neural network for classification. Each layer of this network consists of $1000$ neurons, uses the ReLU activation function for non-linearity, and applies a dropout rate of $0.3$. These configurations are inspired by the model design used for the MNIST dataset.

From the experiments conducted on the benchmark datasets, DiffGrad is found to be the most competitive optimizer. Therefore, for the sake of clarity and focused comparison on real-world datasets, we compare the proposed IAdaPID-ADG optimizer exclusively with DiffGrad.

\subsubsection{IARC Dataset}\label{subsubsec: iarc}
The IARC \cite{IARC2024} dataset is categorized based upon the Transformation Zone (TZ), which refers to the region around the cervical opening where squamous epithelial cells replace columnar epithelial cells. The TZ is typically classified into three categories, namely Type1, Type2, and Type3. The dataset originally contains $318$ images of Type1, $106$ images of Type2, and $147$ images of Type3. We preprocess this dataset following the same procedure described in \cite{saini2025}, resulting in a total of $1575$ images for each type.

\begin{figure*}[!htbp]
	\centering
	
	\makebox[0.33\textwidth][c]{\textbf{Training Loss}}%
	\makebox[0.33\textwidth][c]{\textbf{Training Accuracy}}%
	\makebox[0.33\textwidth][c]{\textbf{Testing Accuracy}}\\[0.5em]
	
	\subfloat[ResNet50]{%
		\includegraphics[width=0.33\textwidth]{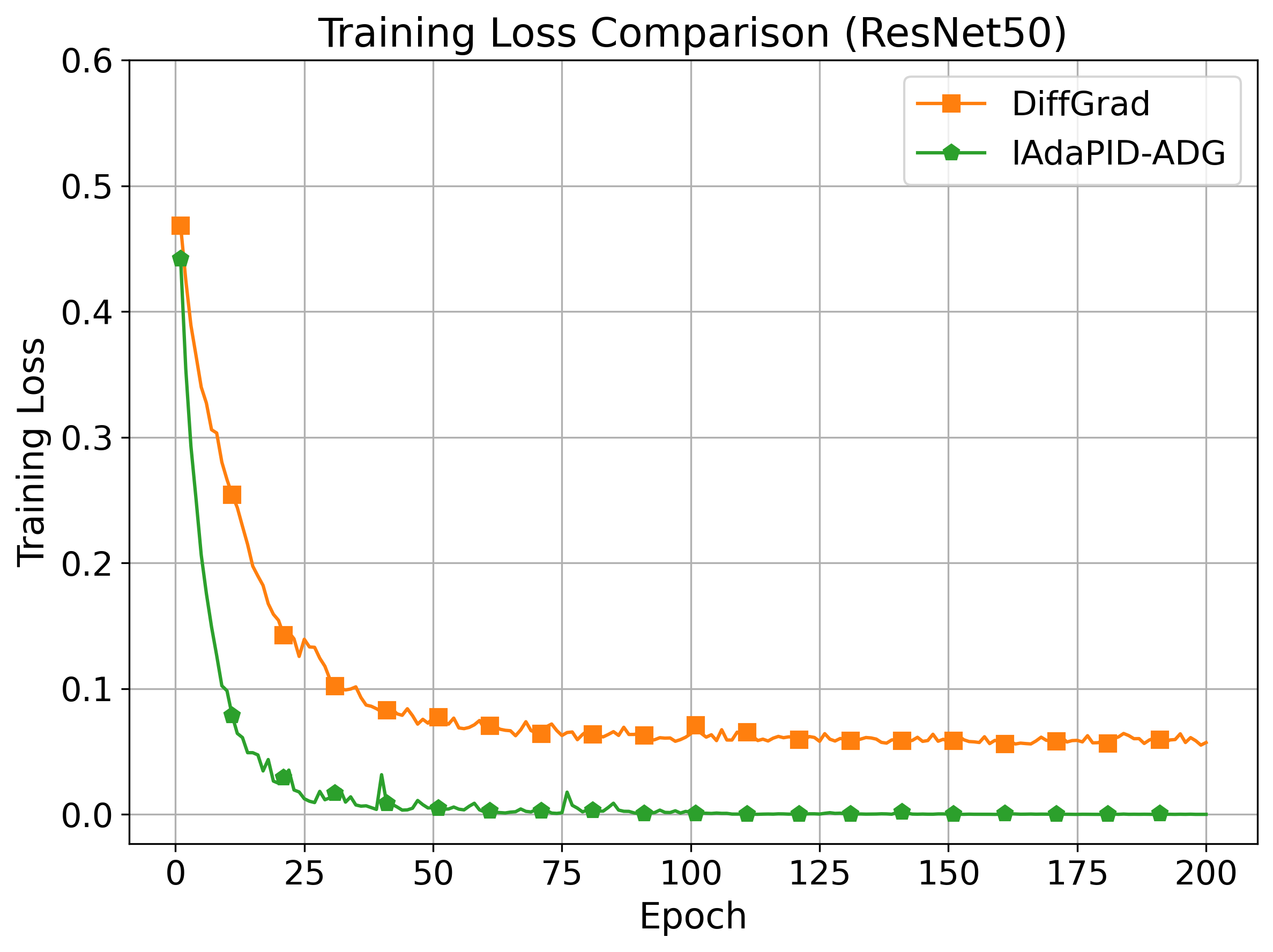}%
	}
	\hfill
	\subfloat[ResNet50]{%
		\includegraphics[width=0.33\textwidth]{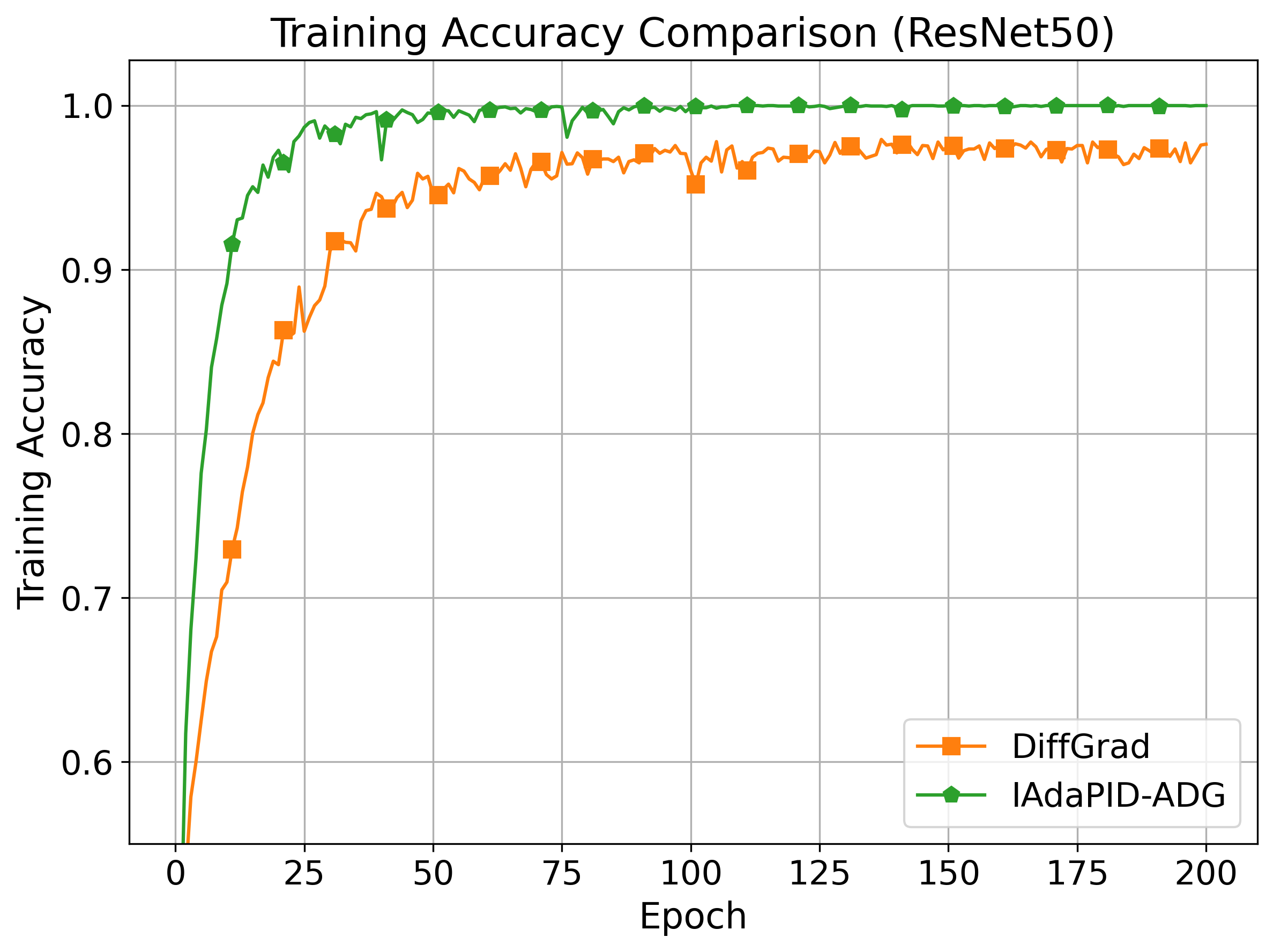}%
	}
	\hfill
	\subfloat[ResNet50]{%
		\includegraphics[width=0.33\textwidth]{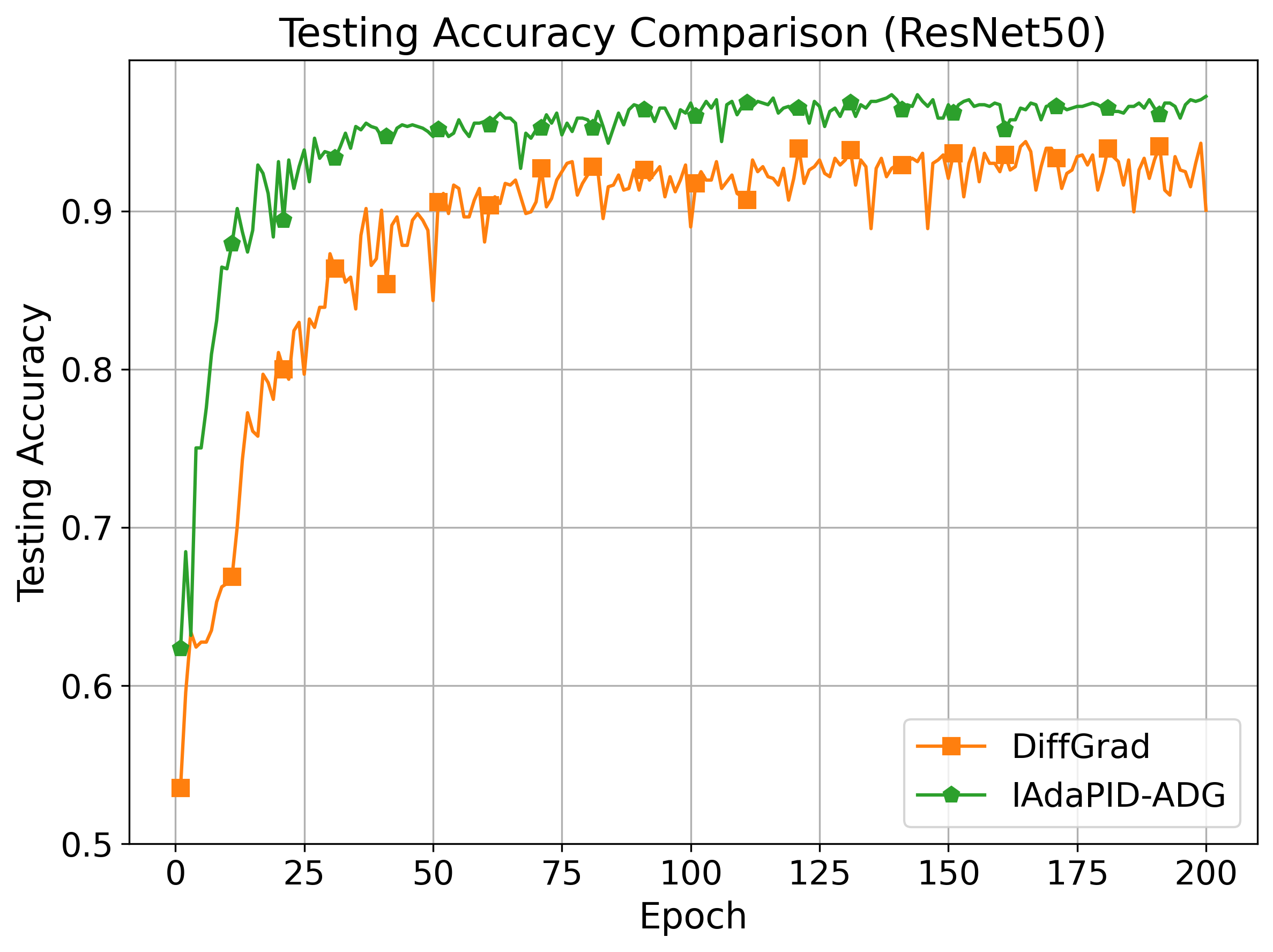}%
	}\\[0.5em]
	
	\subfloat[ResNet101]{%
		\includegraphics[width=0.33\textwidth]{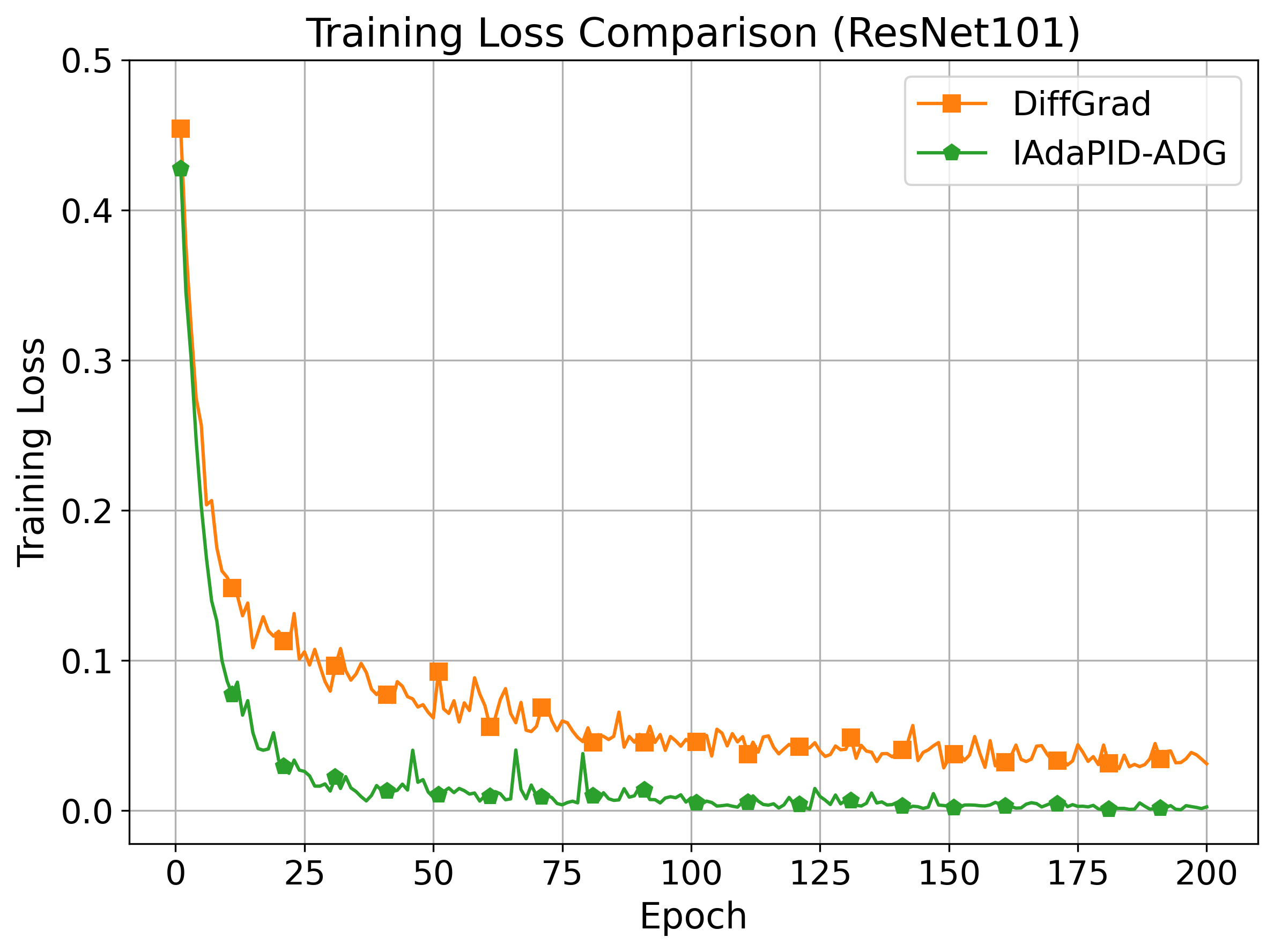}%
	}
	\hfill
	\subfloat[ResNet101]{%
		\includegraphics[width=0.33\textwidth]{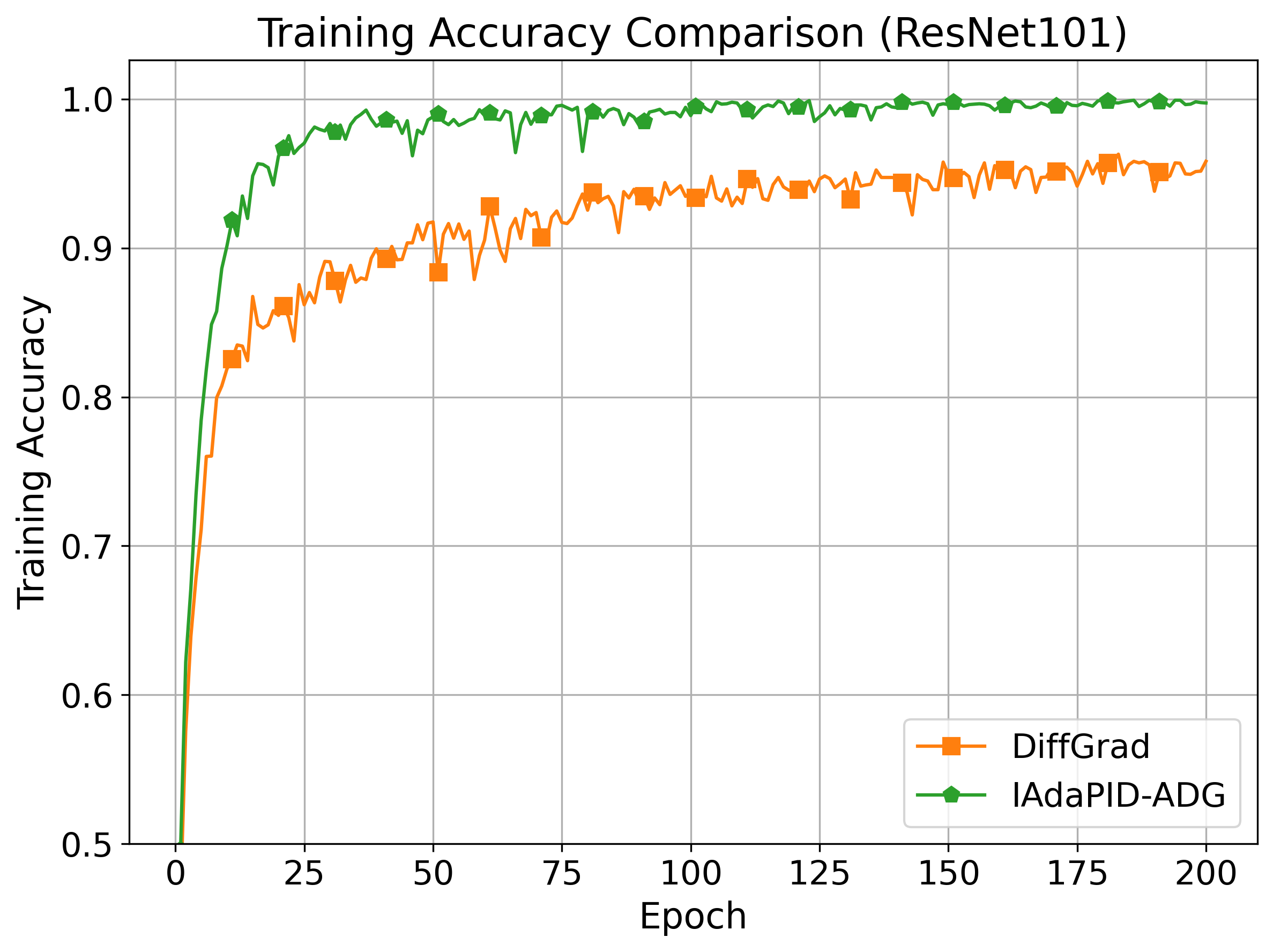}%
	}
	\hfill
	\subfloat[ResNet101]{%
		\includegraphics[width=0.33\textwidth]{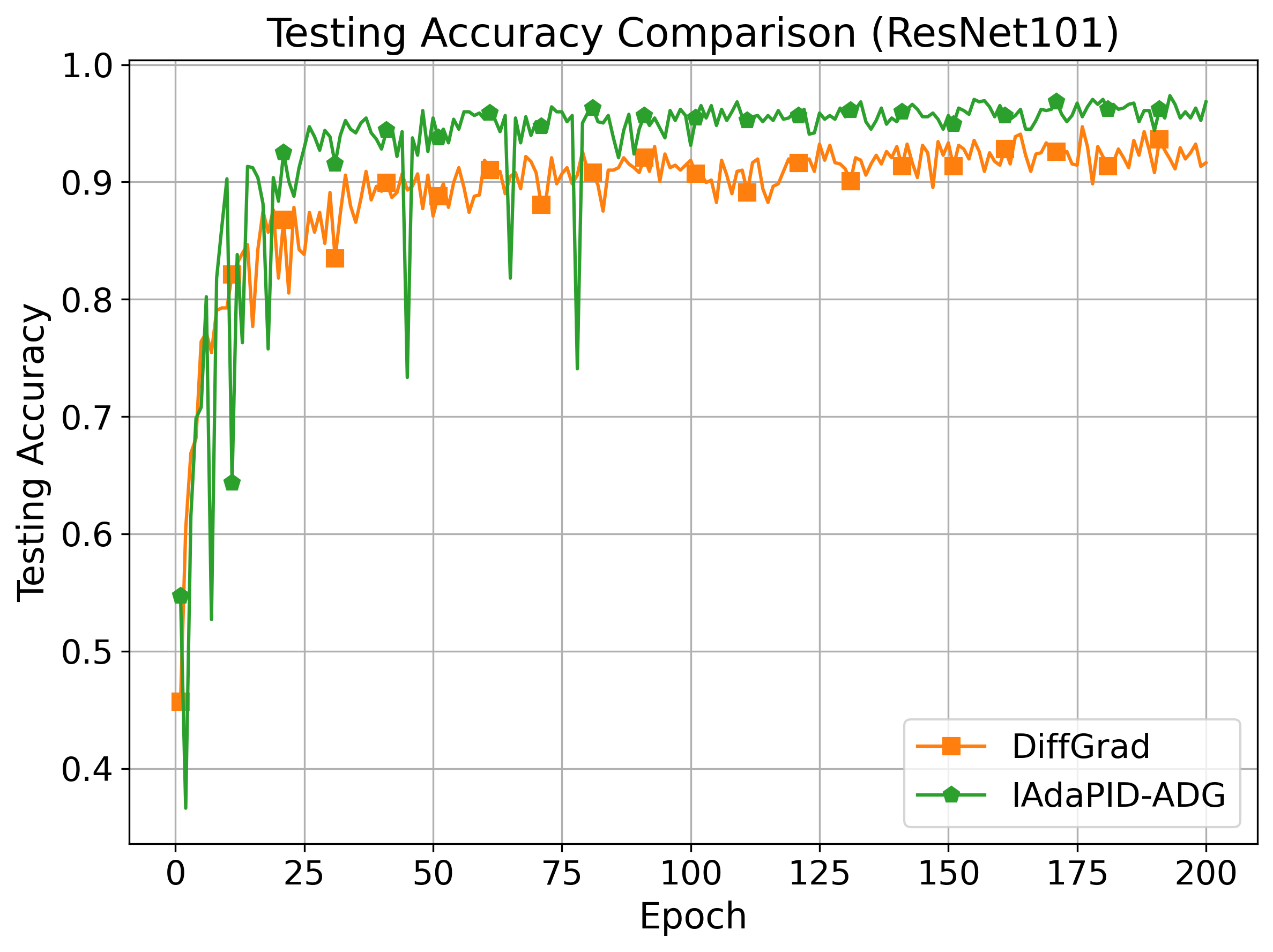}%
	}\\[0.5em]
	
	\subfloat[ResNet152]{%
		\includegraphics[width=0.33\textwidth]{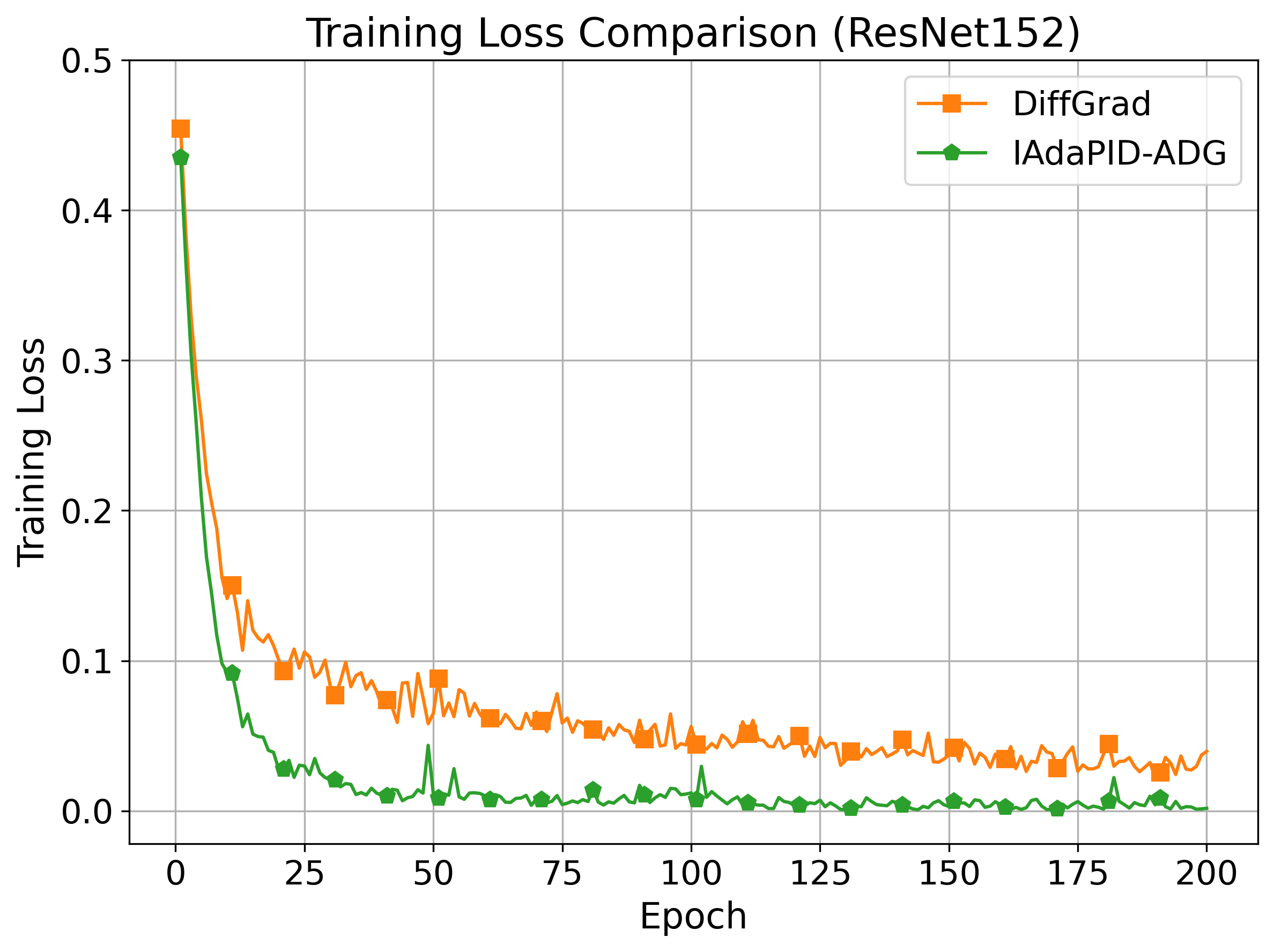}%
	}
	\hfill
	\subfloat[ResNet152]{%
		\includegraphics[width=0.33\textwidth]{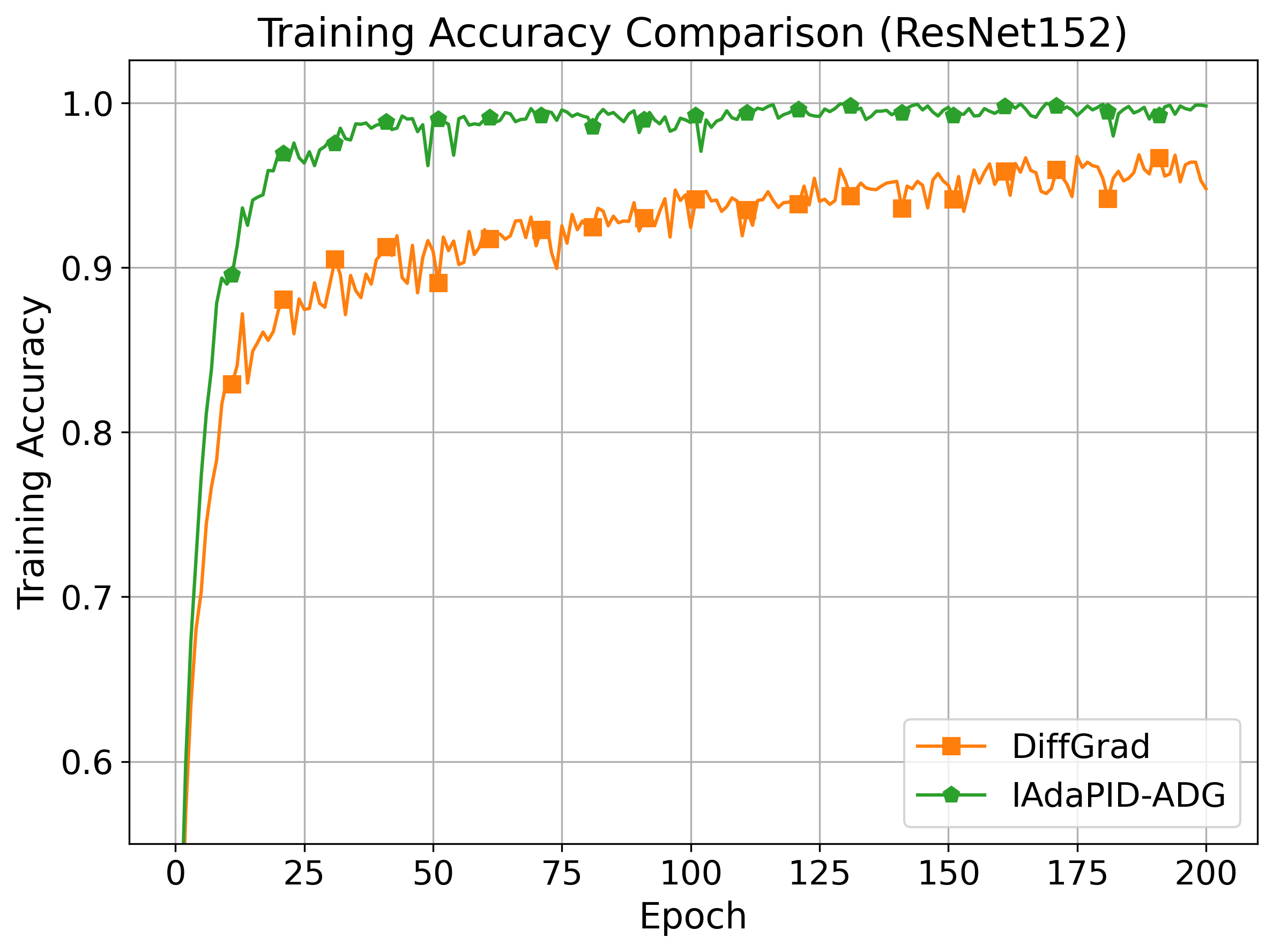}%
	}
	\hfill
	\subfloat[ResNet152]{%
		\includegraphics[width=0.33\textwidth]{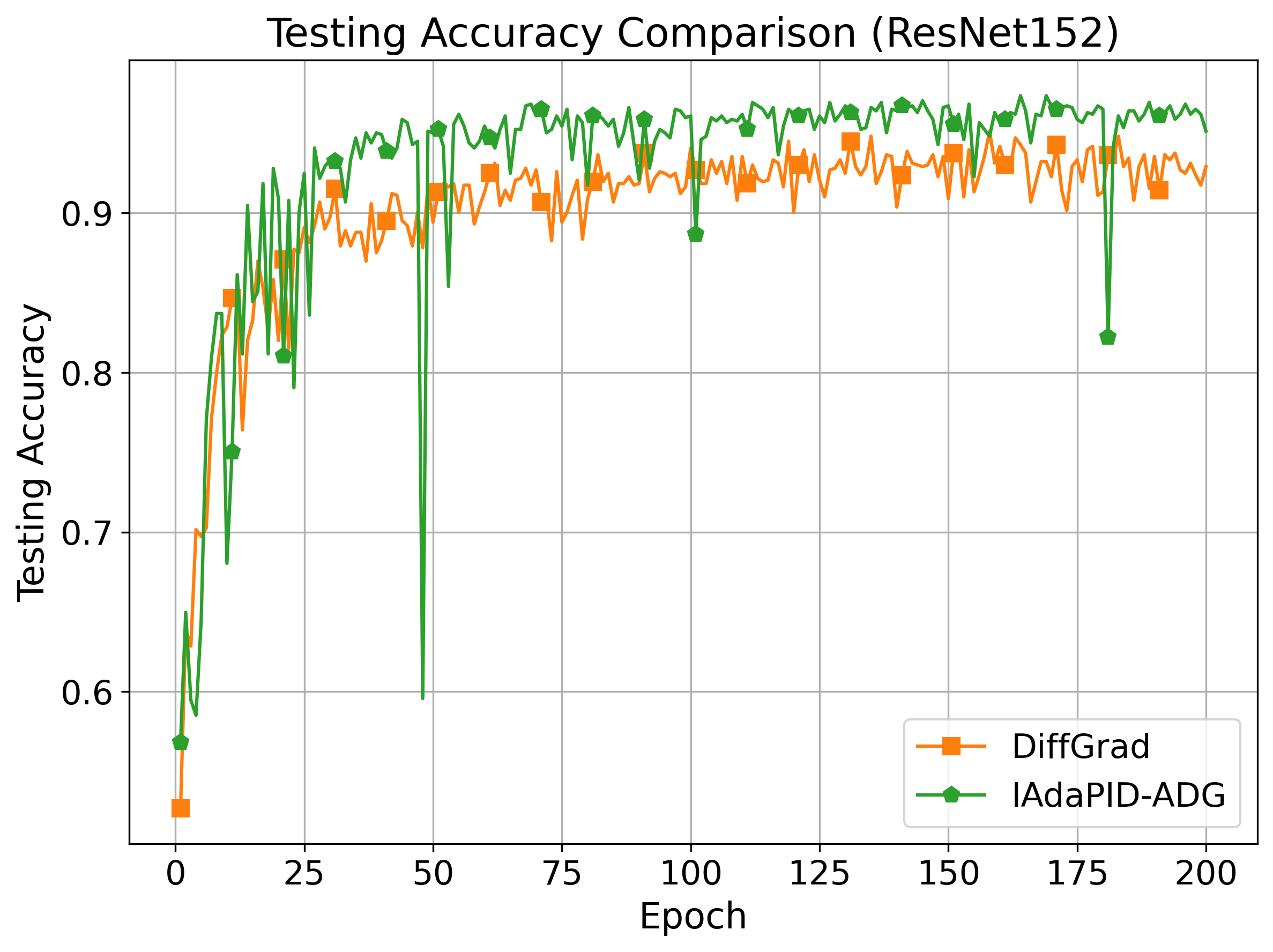}%
	}
	
	\caption{Comparison of DiffGrad and IAdaPID-ADG optimizers on the IARC dataset using (a–c) ResNet50, (d–f) ResNet101, and (g–i) ResNet152. 
		The first column shows training loss, the second shows training accuracy, and the third shows testing accuracy.}
	\label{Fig:IARC_1}
\end{figure*}

Table~\ref{Tab:IARC} presents a comparative analysis of the proposed IAdaPID-ADG and the DiffGrad optimizers across three ResNet architectures, with the corresponding learning curves illustrated in Fig.~\ref{Fig:IARC_1}. On average, the proposed IAdaPID-ADG optimizer achieves a trough in training loss of approximately $0.00048$, while attaining peak training and testing accuracies of $99.95\%$ and $97.35\%$, respectively. In contrast, DiffGrad exhibits comparatively higher training loss and lower training and testing accuracies across all evaluated models. As observed from Fig.~\ref{Fig:IARC_1}, the proposed optimizer consistently maintains a lower training loss curve and higher training and testing accuracy curves throughout the training process, indicating faster convergence and improved stability, whereas DiffGrad shows relatively slower convergence with more fluctuations.

\renewcommand{\arraystretch}{1.3}
\begin{table}[!htbp]
	\caption{Comparison of DiffGrad and IAdaPID-ADG optimizers on the IARC dataset across three ResNet models.}
	\centering
	\resizebox{\columnwidth}{!}{ 
		\begin{tabular}{l|l|l|l|l}
			\hline 
			\multirow{2}{*}{Model} & \multirow{2}{*}{Optimizer} & Training & Training & Testing \\ 
			& & Loss & Accuracy (\%) & Accuracy (\%) \\
			\hline
			
			\multirow{2}{*}{ResNet50} 
			& DiffGrad & 0.05720 & 97.93 & 94.07 \\
			& \textbf{IAdaPID-ADG} & \textbf{0.00004} & \textbf{100} & \textbf{97.35} \\ 
			\hline
			
			\multirow{2}{*}{ResNet101} 
			& DiffGrad & 0.02805 & 96.30 & 93.43 \\
			& \textbf{IAdaPID-ADG} & \textbf{0.00064} & \textbf{99.92} & \textbf{97.35} \\ 
			\hline
			
			\multirow{2}{*}{ResNet152} 
			& DiffGrad & 0.02606 & 96.90 & 94.17 \\
			& \textbf{IAdaPID-ADG} & \textbf{0.00077} & \textbf{99.92} & \textbf{97.35} \\ 
			\hline
		\end{tabular}
	}
	\label{Tab:IARC}
\end{table}

\subsubsection{AnnoCerv Dataset}\label{subsubsec: anno}
The AnnoCerv \cite{Minciuna2023} dataset is categorized into three categories based on their scores, which range from $0$ to $10$. Images with scores from $0$ to $4$ fall into Category1 (low-grade/CIN1), those with scores from $5$ to $6$ fall into Category2 (high-grade/CIN2+), and those with scores from $7$ to $10$ belong to Category3 (high-grade/suspected invasive cancer/CIN2+). The dataset originally contains $311$ images of Category1, $124$ images of Category2, and $96$ images of Category3. We again apply the same preprocessing methods described in \cite{saini2025}, resulting in a total of $1555$ images for each category.

\begin{figure*}[!htbp]
	\centering
	
	\makebox[0.33\textwidth][c]{\textbf{Training Loss}}%
	\makebox[0.33\textwidth][c]{\textbf{Training Accuracy}}%
	\makebox[0.33\textwidth][c]{\textbf{Testing Accuracy}}\\[0.5em]
	
\subfloat[ResNet50]{%
	\includegraphics[width=0.33\textwidth]{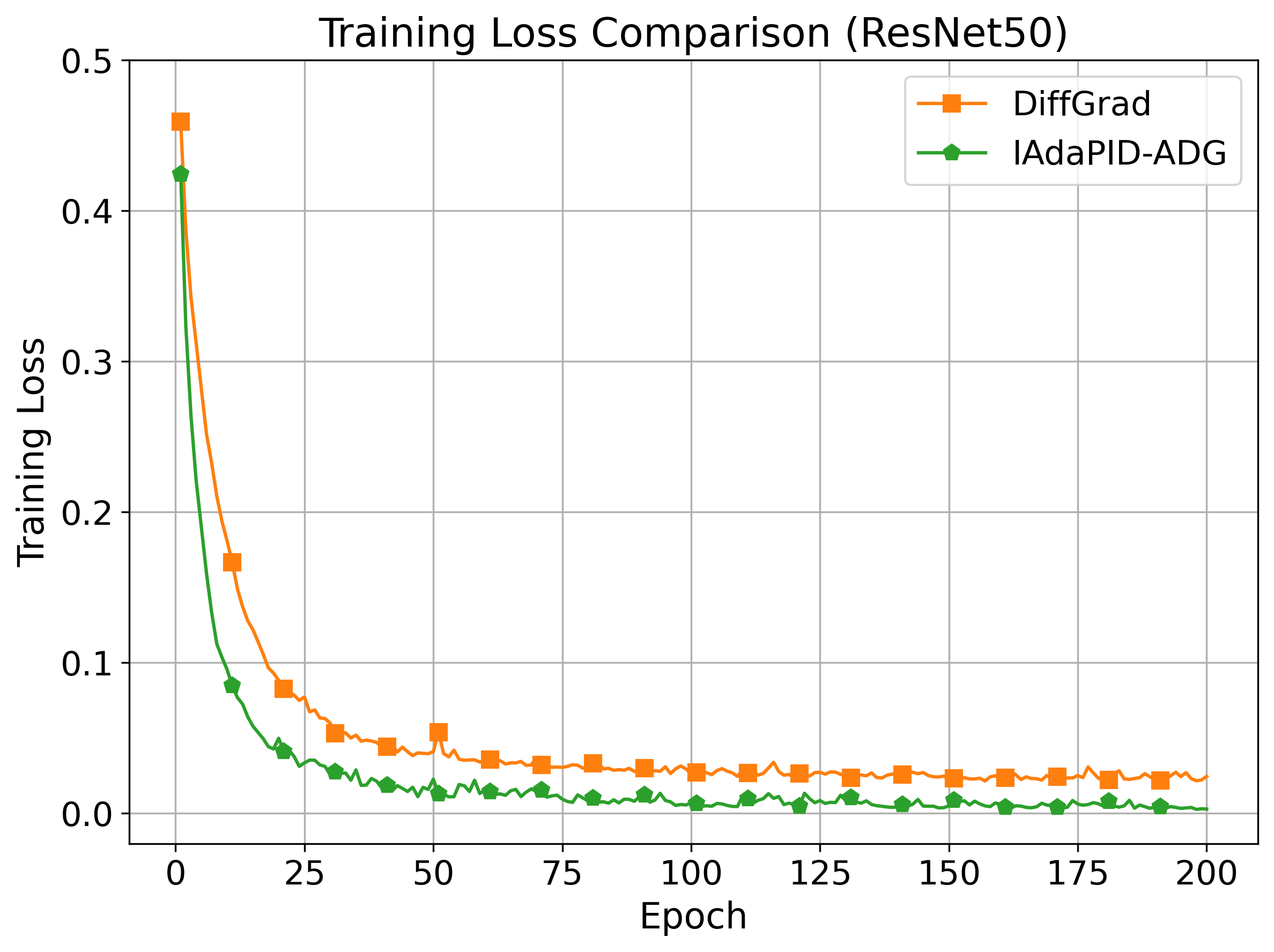}%
}
\hfill
\subfloat[ResNet50]{%
	\includegraphics[width=0.33\textwidth]{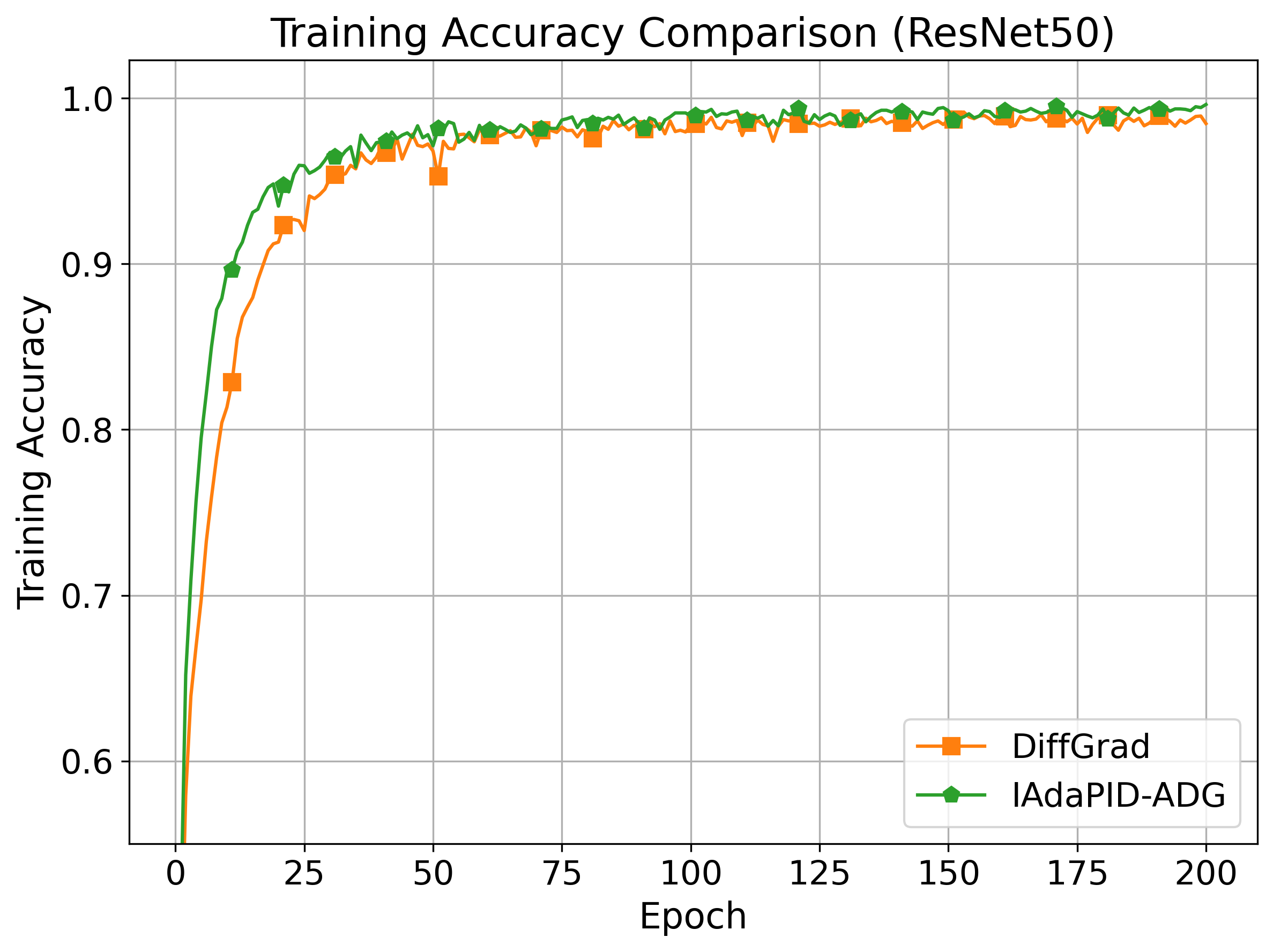}%
}
\hfill
\subfloat[ResNet50]{%
	\includegraphics[width=0.33\textwidth]{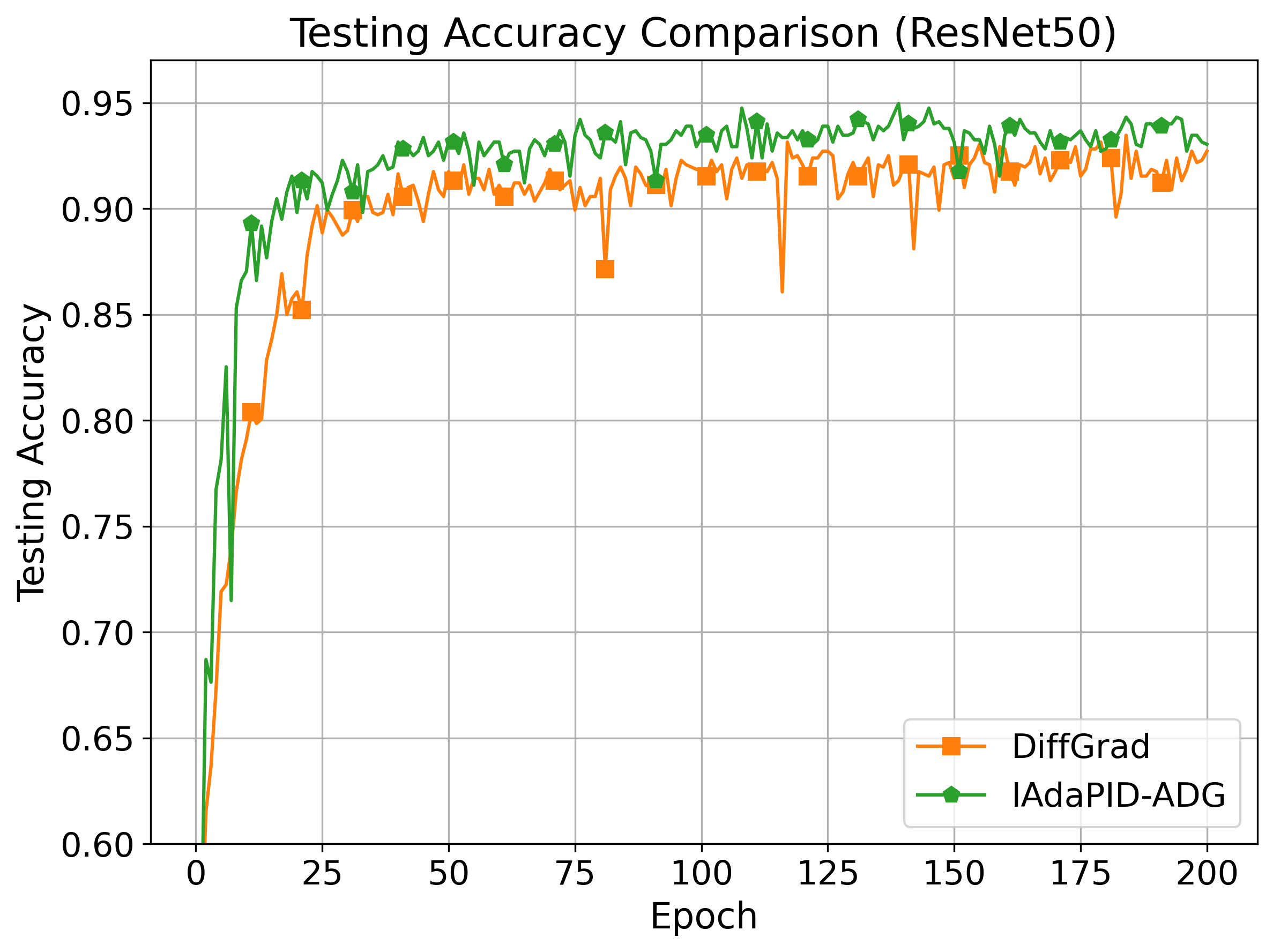}%
}\\[0.5em]

\subfloat[ResNet101]{%
	\includegraphics[width=0.33\textwidth]{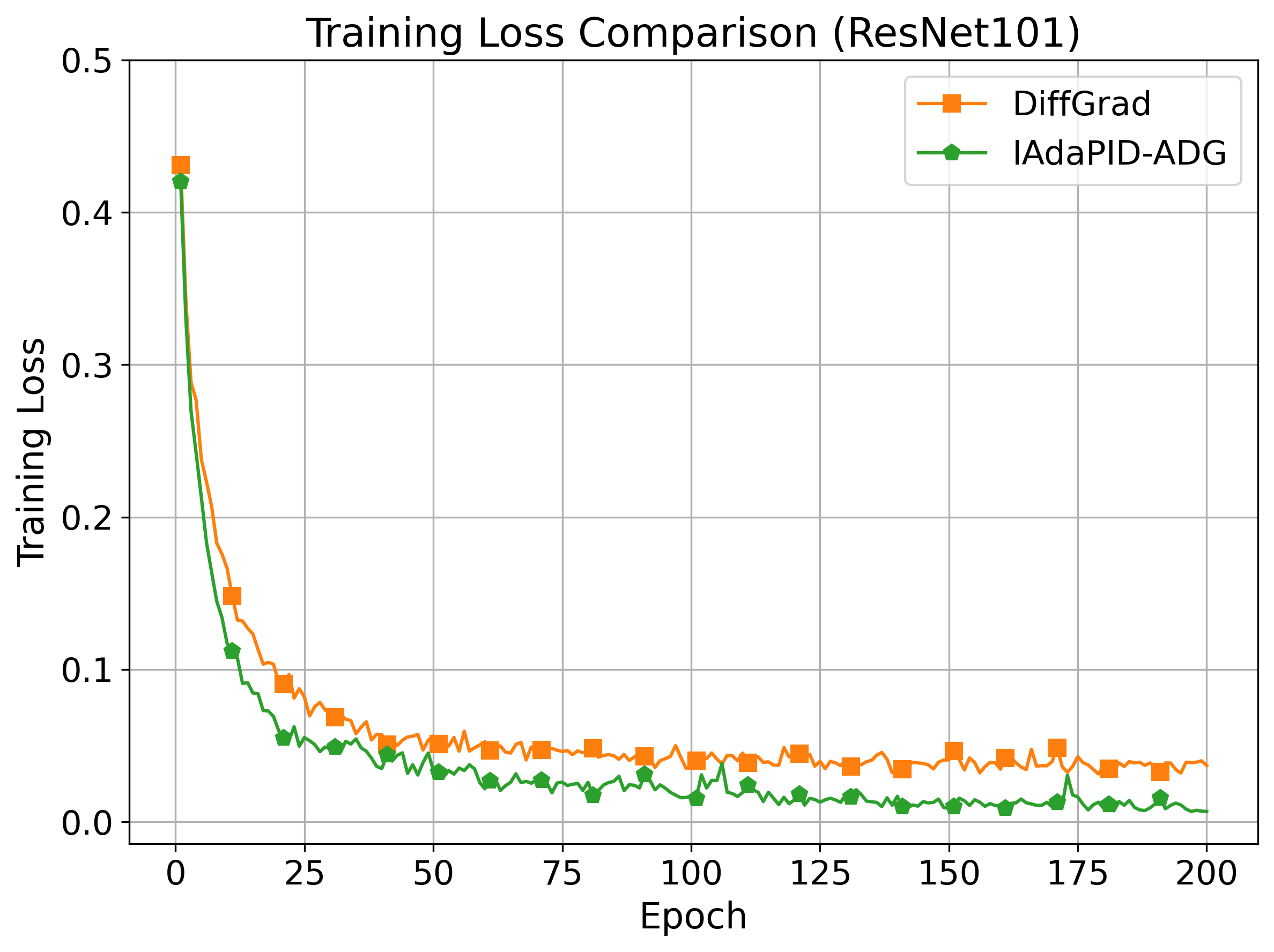}%
}
\hfill
\subfloat[ResNet101]{%
	\includegraphics[width=0.33\textwidth]{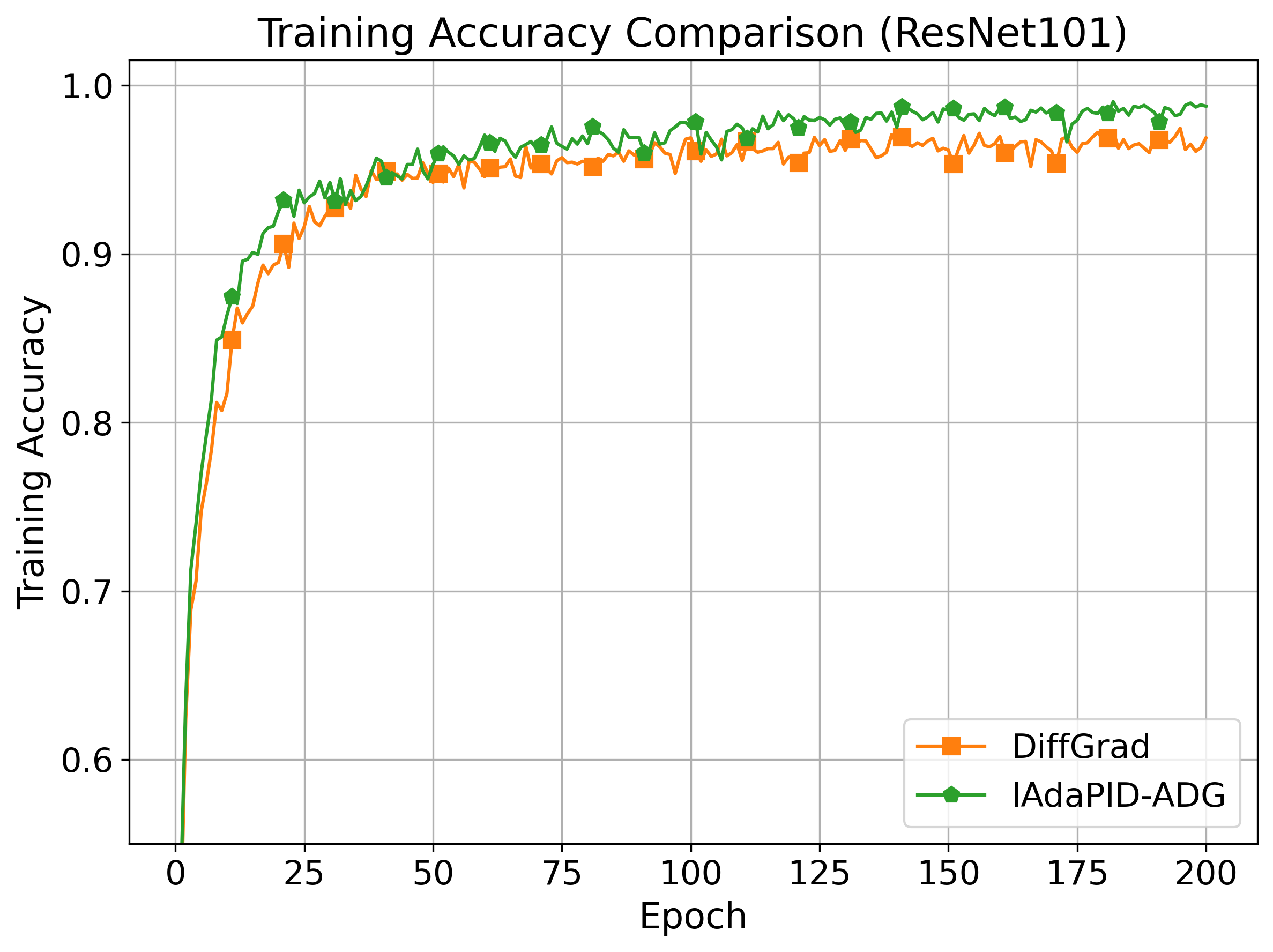}%
}
\hfill
\subfloat[ResNet101]{%
	\includegraphics[width=0.33\textwidth]{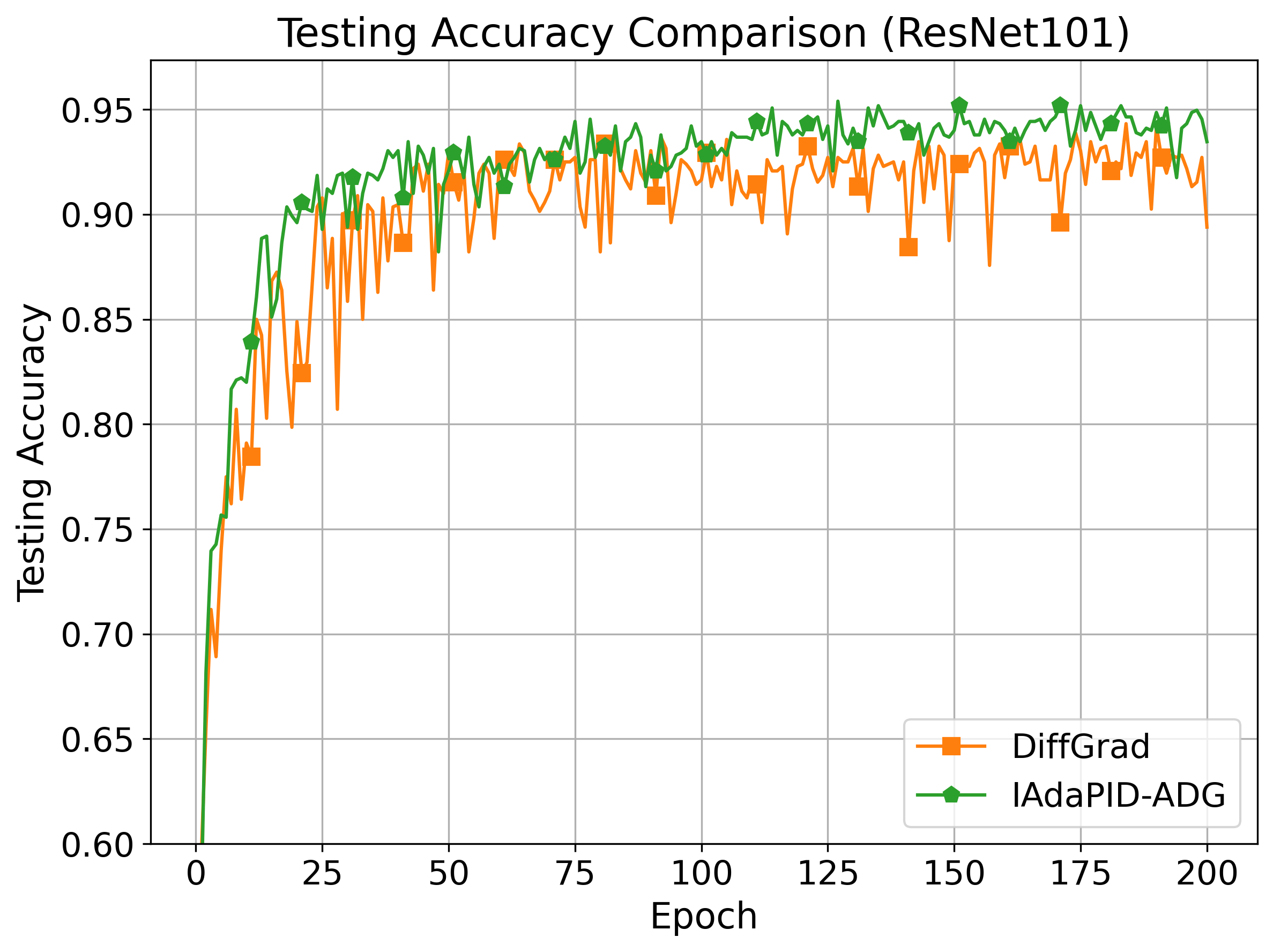}%
}\\[0.5em]

\subfloat[ResNet152]{%
	\includegraphics[width=0.33\textwidth]{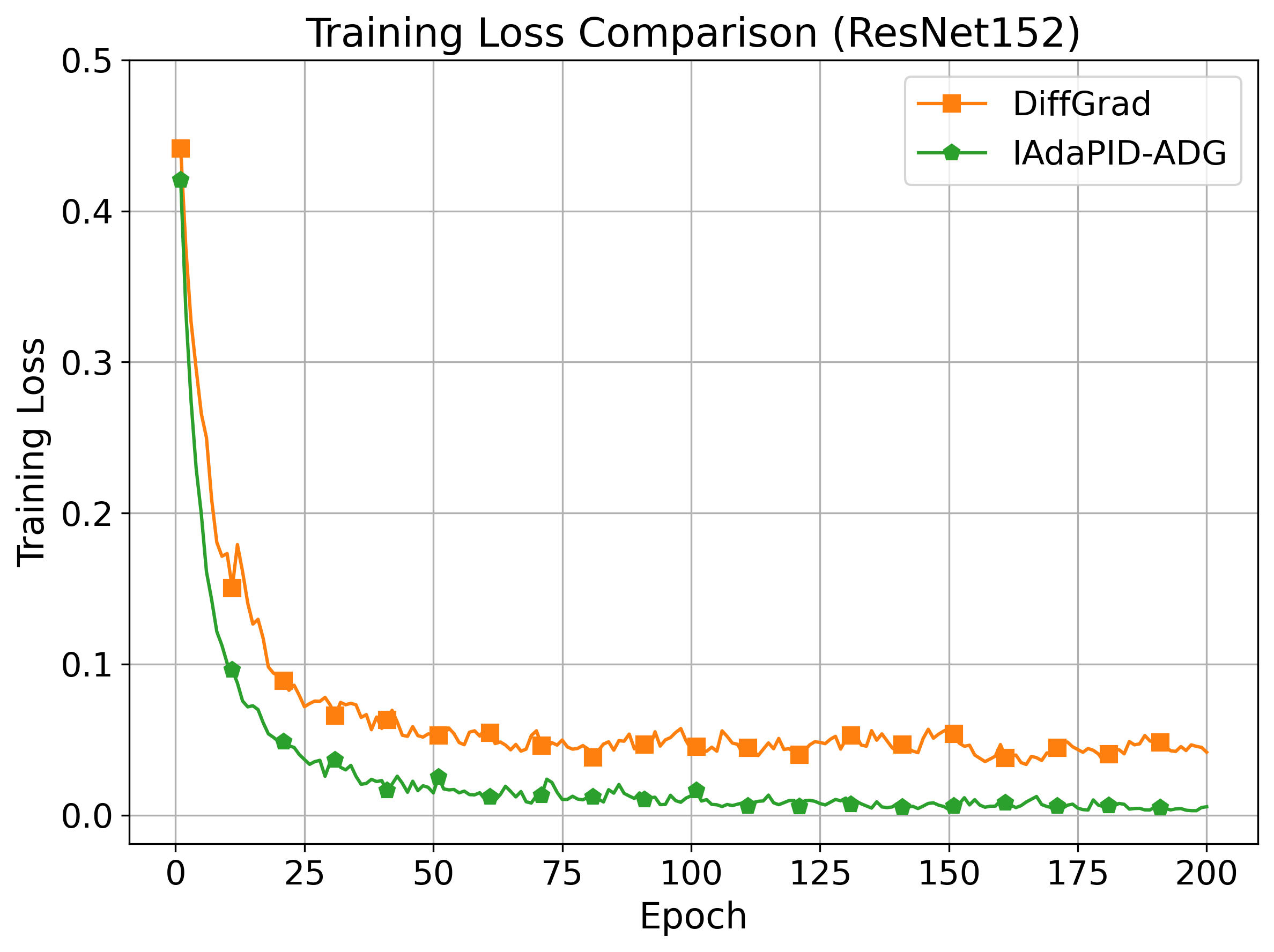}%
}
\hfill
\subfloat[ResNet152]{%
	\includegraphics[width=0.33\textwidth]{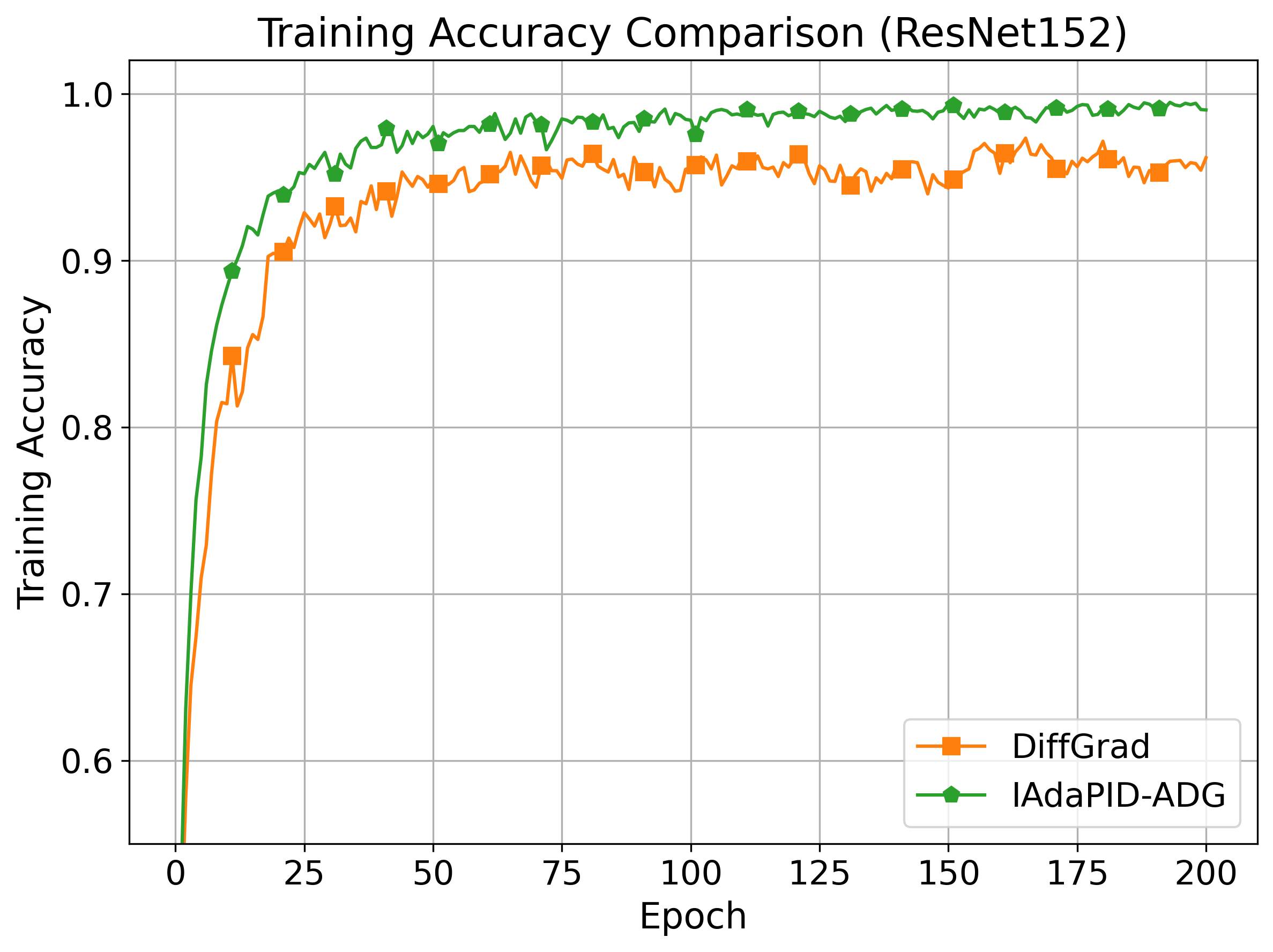}%
}
\hfill
\subfloat[ResNet152]{%
	\includegraphics[width=0.33\textwidth]{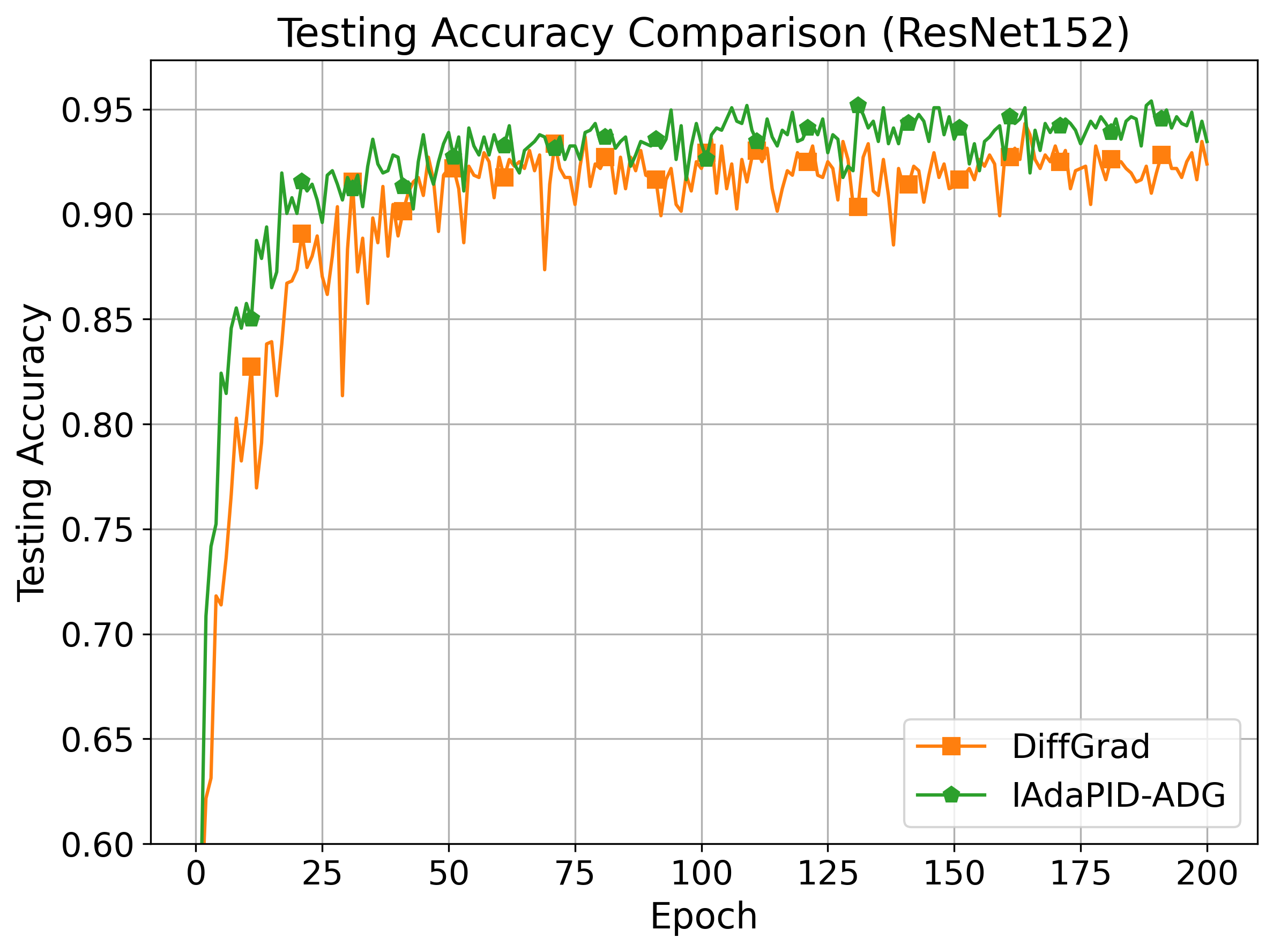}%
}
	
	\caption{Comparison of DiffGrad and IAdaPID-ADG optimizers on the AnnoCerv dataset using (a–c) ResNet50, (d–f) ResNet101, and (g–i) ResNet152. The first column shows training loss, the second shows training accuracy, and the third shows testing accuracy.}
	\label{Fig:Anno_1}
\end{figure*}

Table~\ref{Tab:AnnoCerv} provides a comparative evaluation of the proposed IAdaPID-ADG and the DiffGrad  optimizers using the same three ResNet architectures, with the corresponding training dynamics illustrated in Fig.~\ref{Fig:Anno_1}. On average, IAdaPID-ADG reaches a trough in training loss of $0.00407$ and achieves peak training and testing accuracies of $99.32\%$ and $95.25\%$, respectively. In contrast, DiffGrad exhibits comparatively higher loss levels and does not attain similar peak accuracy performance across the evaluated models. As depicted in Fig.~\ref{Fig:Anno_1}, the proposed optimizer consistently maintains lower loss curves and higher accuracy trajectories throughout the training process, indicating faster convergence and improved stability. Moreover, the learning curves of IAdaPID-ADG appear smoother with fewer oscillations, whereas DiffGrad shows relatively slower convergence with noticeable fluctuations. 

Overall, these results show that IAdaPID-ADG provides better optimization efficiency and improved generalization across all considered architectures on real-world datasets.

\renewcommand{\arraystretch}{1.3}
\begin{table}[!htbp]
	\caption{Comparison of DiffGrad and IAdaPID-ADG optimizers on the AnnoCerv dataset across three ResNet models.}
	\centering
	\resizebox{\columnwidth}{!}{ 
		\begin{tabular}{l|l|l|l|l}
			\hline 
			\multirow{2}{*}{Model} & \multirow{2}{*}{Optimizer} & Training & Training & Testing \\ 
			& & Loss & Accuracy (\%) & Accuracy (\%) \\
			\hline
			
			\multirow{2}{*}{ResNet50} 
			& DiffGrad & 0.03271 & 97.18 & 93.46 \\
			& \textbf{IAdaPID-ADG} & \textbf{0.00252} & \textbf{99.49} & \textbf{94.96} \\ 
			\hline
			
			\multirow{2}{*}{ResNet101} 
			& DiffGrad & 0.03202 & 97.45 & 94.31 \\
			& \textbf{IAdaPID-ADG} & \textbf{0.00677} & \textbf{99.03} & \textbf{95.39} \\ 
			\hline
			
			\multirow{2}{*}{ResNet152} 
			& DiffGrad & 0.03357 & 97.34 & 94.31 \\
			& \textbf{IAdaPID-ADG} & \textbf{0.00293} & \textbf{99.44} & \textbf{95.39} \\ 
			\hline
		\end{tabular}
	}
	\label{Tab:AnnoCerv}
\end{table}

\section{Ablation Study}\label{sec:ablation}
This section presents the ablation study of our optimizer. As discussed earlier, the proposed optimizer combines the principles of AMSGrad and DiffGrad with AdaPID. Therefore, we perform an ablation study to analyze the individual contribution of each of these components to the overall performance. Without loss of generality we perform this ablation study on the MNIST dataset. The underlying model and hyperparameter configurations are the same as those used earlier for the MNIST experiments.

\begin{figure*}[!htbp]
	\centering
	\subfloat[]{%
		\includegraphics[width=0.33\textwidth]{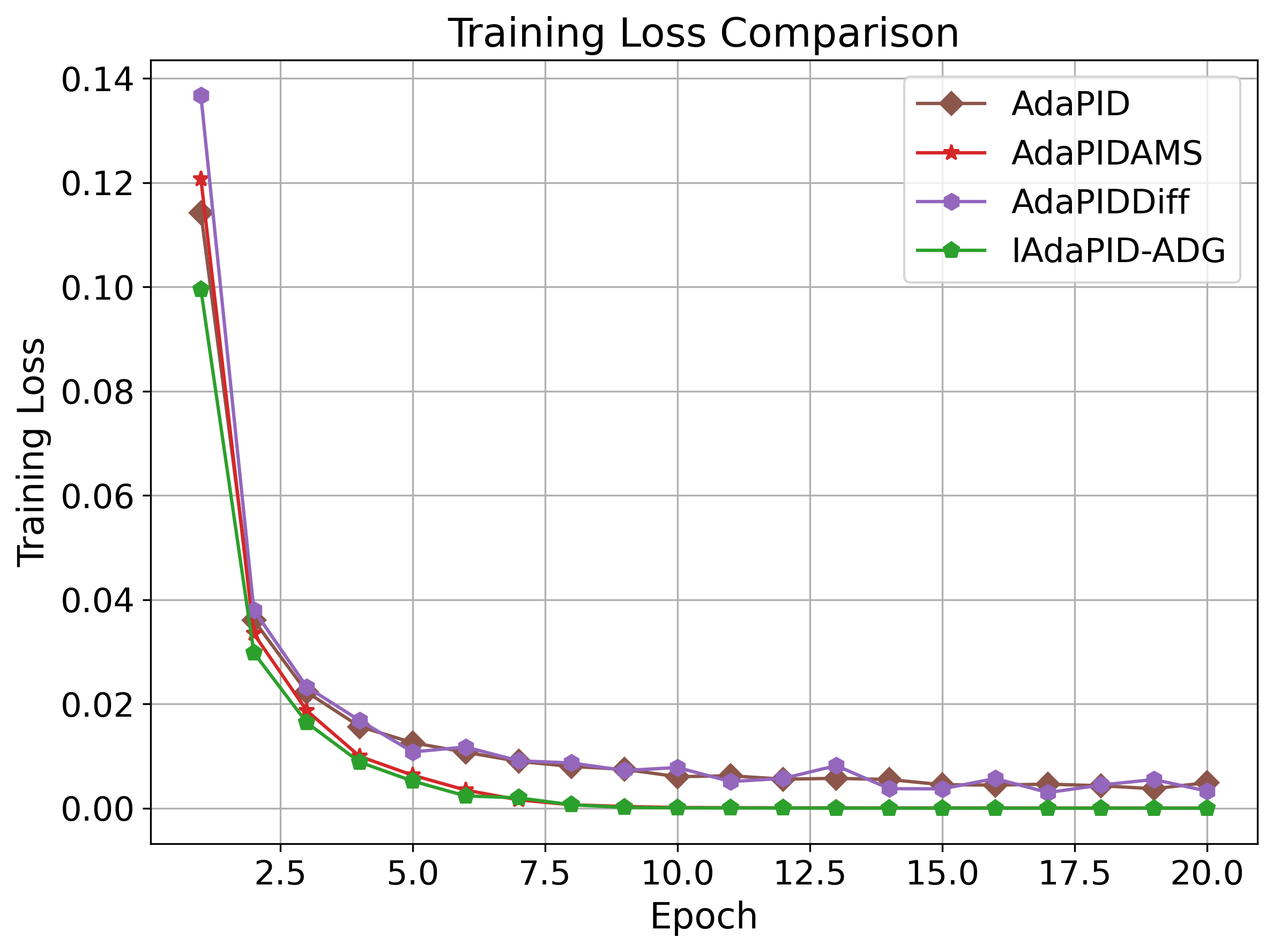}%
	}
	\hfill
	\subfloat[]{%
		\includegraphics[width=0.33\textwidth]{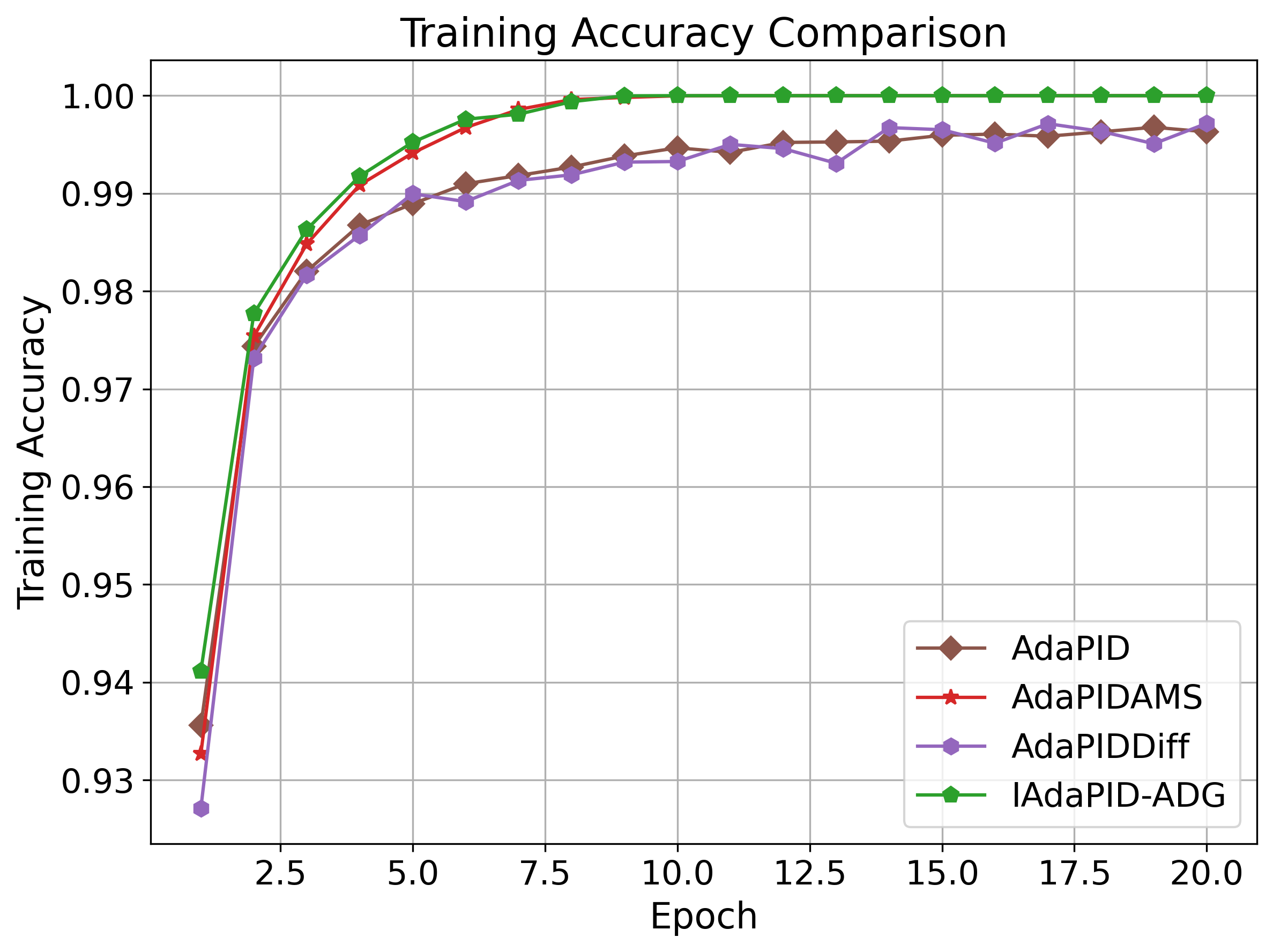}%
	}
	\hfill
	\subfloat[]{%
		\includegraphics[width=0.33\textwidth]{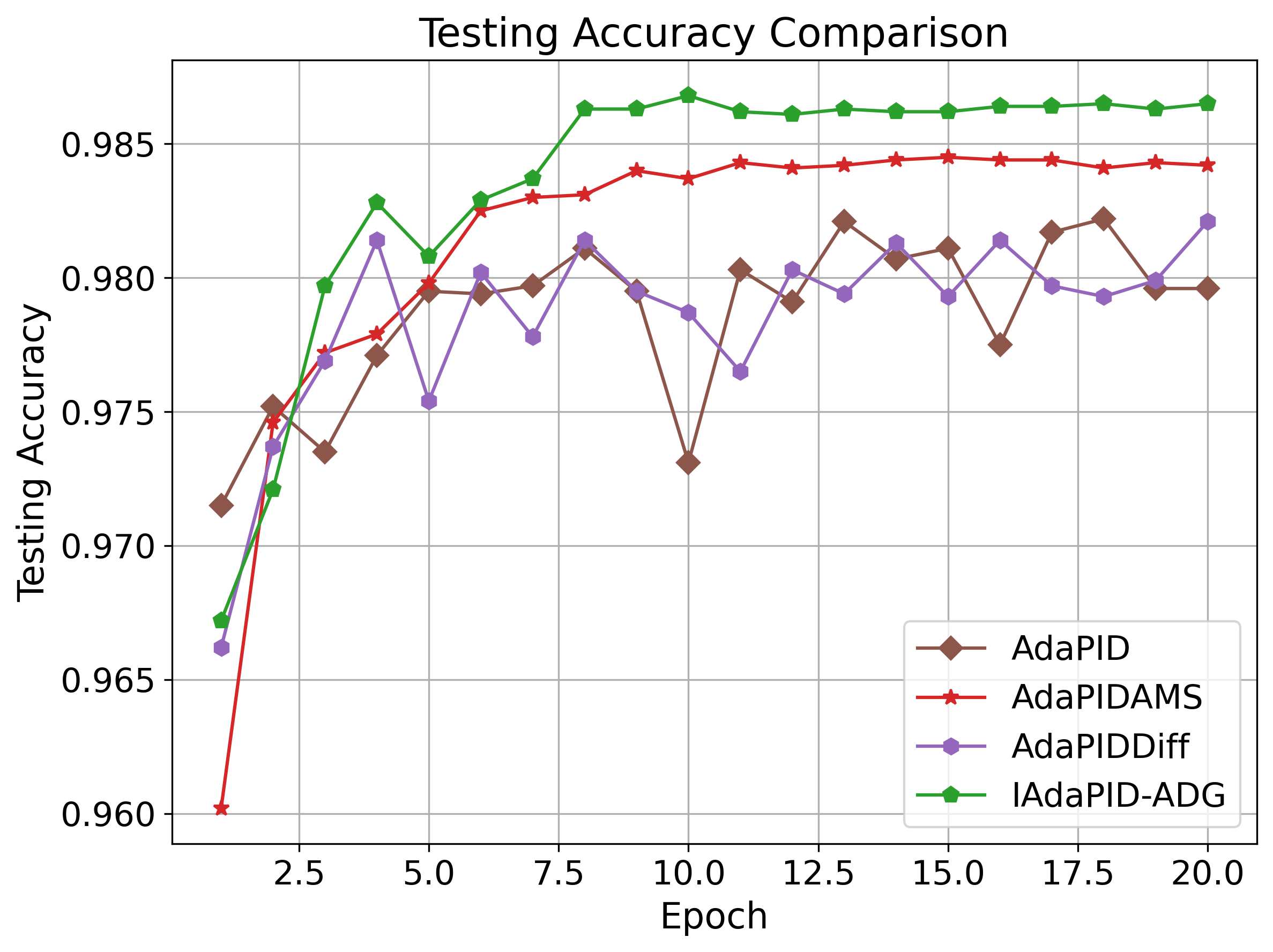}%
	}
	\caption{Comparison of AdaPID, AdaPIDAMS, AdaPIDDiff, and IAdaPID-ADG optimizers on the MNIST dataset; (a) training loss, (b) training accuracy, and (c) testing accuracy.}
	\label{Fig: Ablation}
\end{figure*}

Table~\ref{Tab: Ablation} shows the results of the ablation study, along with the learning curves in Fig.~\ref{Fig: Ablation}. From Table~\ref{Tab: Ablation}, the proposed IAdaPID-ADG optimizer gives the best overall performance among all configurations. It reaches a trough in training loss of $0.00002$ and achieves peak training and testing accuracies of $100\%$ and $98.68\%$, respectively. In comparison, the baseline AdaPID and its improved versions, where components are added step by step, have higher training loss and do not reach similar peak accuracy. From Fig.~\ref{Fig: Ablation}, it can be seen that the proposed optimizer keeps the training loss lower and the accuracy higher throughout training, showing faster learning and better stability. These results indicate that each added component improves performance, and combining them gives the best overall results.

\renewcommand{\arraystretch}{1.3}
\begin{table}[!htbp]
	\caption{Comparison of AdaPID, AdaPIDAMS, AdaPIDDiff, and IAdaPID-ADG optimizers on the MNIST dataset.}
	\centering
	\scriptsize
	\resizebox{0.95\columnwidth}{!}{   
		\begin{tabular}{l|l|l|l}
			\hline 
			\multirow{2}{*}{Optimizer} & \multicolumn{2}{c|}{Training} & Testing \\ \cline{2-3}
			& Loss & Accuracy (\%)& Accuracy (\%)\\
			\hline
			
			AdaPID    &  0.004 & 99.67 & 98.11 \\ 
			AdaPIDDiff   & 0.003& 99.70 &98.14  \\ 
			AdaPIDAMS         &  0.00004& 100 & 98.30\\
			
			\textbf{IAdaPID-ADG} & \textbf{0.00002} &  \textbf{100}& \textbf{98.68} \\ 
			\hline
		\end{tabular}
	}
	\label{Tab: Ablation}
\end{table}

\section{Conclusion}\label{sec:conclusion}
In this work, we propose a novel optimizer called IAdaPID-ADG to address two critical limitations inherited by AdaPID from Adam, namely convergence and stability issues. \textit{First}, the concept of a non-increasing effective learning rate, introduced in AMSGrad, is uniquely integrated into AdaPID to address the convergence issue. \textit{Second}, a gradient-difference-based modulation factor, inspired by DiffGrad, is innovatively incorporated into the optimizer obtained from the first step to fix the stability issue.

The proposed optimizer is extensively evaluated on four different datasets to assess its generalizability. The first two datasets are standard benchmark datasets, namely MNIST and CIFAR10, while the remaining two are real-world cervical cancer datasets, namely IARC and AnnoCerv. Across all datasets, our proposed optimizer achieves training losses that are 2--3 orders of magnitude lower than those of the competing optimizers. IAdaPID-ADG attains training accuracy close to $100\%$ and consistently performs substantially better than the existing optimizers on all datasets. Furthermore, the proposed optimizer achieves better testing accuracy than the competing optimizers across all datasets. On the benchmark datasets, the testing accuracy ranges from $81\%$ to $99\%$, with improvements of $0.6\%$ to $2.5\%$. On the real-world datasets, our optimizer achieves testing accuracy ranging from $95\%$ to $97\%$, with gains of $1.08\%$ to $3.7\%$.

An ablation analysis is also performed on the MNIST dataset to examine the effect of each individual component of IAdaPID-ADG. The study confirms that the integration of both the AMSGrad-based learning rate strategy and the DiffGrad-inspired modulation mechanism plays an important role in enhancing the overall performance.

As future work, two promising directions are identified. First, the proposed IAdaPID-ADG optimizer can be applied to train models in complex information systems, such as recommendation systems, where effective optimization plays a key role \cite{murthy2006simpel, kim2005Explain}. Second, the convergence properties of IAdaPID-ADG can be further analyzed through the lens of iterative numerical methods, such as Krylov subspace based approaches, which offer a rich theoretical framework \cite{ahuja2011recycling, choudhary2018stability}.



\vspace{0.5cm}
\noindent\textbf{Code \& Data Availability:} The code and data are  available \href{https://drive.google.com/drive/folders/1LoQsDqY9KHq77ZzdlflOalC6enTc-13w?usp=drive_link}{here}.

\bibliographystyle{IEEEtran}
\bibliography{bibliography}

\end{document}